%% file: oneclass-iccv19.tex
\documentclass[10pt,twocolumn,letterpaper]{article}

\usepackage{iccv}
\usepackage{amsmath}
\usepackage{amsthm}
\usepackage{amssymb}
\usepackage{times}  %Required
\usepackage{bm}
\usepackage{helvet}  %Required
\usepackage{courier}  %Required
\usepackage{url}  %Required
\usepackage{graphicx}  %Required
\usepackage{color}
\usepackage{adjustbox,lipsum}
\usepackage{subfigure}

\input{definitions.tex}

\usepackage[pagebackref=true,breaklinks=true,letterpaper=true,colorlinks,bookmarks=false]{hyperref}

\iccvfinalcopy % *** Uncomment this line for the final submission

\def\iccvPaperID{943} % *** Enter the ICCV Paper ID here

\ificcvfinal\fi
\begin{document}

%%%%%%%%% TITLE
\title{GODS: Generalized One-class Discriminative Subspaces for Anomaly Detection}

\author{Jue Wang\thanks{Work done while interning at MERL.}\\
Australian National University, Canberra\\
{\tt\small jue.wang@anu.edu.au}
\and
Anoop Cherian\\
Mitsubishi Electric Research Labs, Cambridge, MA\\
{\tt\small cherian@merl.com}
}

\maketitle%Required

\input{abstract}
\input{intro}

\input{background}
\input{algo}
\input{expts}

\input{conclude}
%\newpage
%\input{suppl}

\bibliographystyle{ieee}
\bibliography{one-class-bib}

\end{document}

%% file: definitions.tex
\newcommand{\vx}{\mathbf{x}}
\newcommand{\vw}{\mathbf{w}}
\newcommand{\vc}{\mathbf{c}}
\newcommand{\vb}{\mathbf{b}}

\newcommand{\valpha}{\bm{\alpha}}
\newcommand{\vOmega}{\bm{\Omega}}

\newcommand{\half}{\frac{1}{2}}
\newcommand{\reals}[1]{\mathbb{R}^{#1}}
\newcommand{\mW}{\mathbf{W}}
\newcommand{\mZ}{\mathbf{Z}}

\newcommand{\stiefel}[1]{\mathcal{S}^{#1}_d}
\newcommand{\dataset}{\mathcal{D}}
\newcommand{\dl}{\mathcal{D}_o}
\newcommand{\dbar}{\mathcal{\overline{D}}}
\newcommand{\subjectto}{\text{ s.t. }}
\newcommand{\unitsphere}[1]{U^{#1}}
\newcommand{\eye}[1]{\mathbf{I}_{#1}}

\newcommand{\enorm}[1]{\left\|{#1}\right\|_2}
\newcommand{\fnorm}[1]{\left\|{#1}\right\|_F}

\newcommand{\set}[1]{\left\{#1\right\}}
\newcommand{\hinge}[1]{\left[{#1}\right]_+}

\DeclareMathOperator*{\sign}{sgn}
\DeclareMathOperator*{\dist}{dist^2}
\DeclareMathOperator*{\Dist}{dist}

\DeclareMathOperator*{\distW}{dist_W^2}
\DeclareMathOperator*{\DistW}{dist_W}
\DeclareMathOperator*{\argmax}{arg\,max}
\DeclareMathOperator*{\argmin}{arg\,min}
\DeclareMathOperator*{\grad}{grad}

%% file: abstract.tex
\begin{abstract}
One-class learning is the classic problem of fitting a model to data for which annotations are available only for a single class. In this paper, we propose a novel objective for one-class learning. Our key idea is to use a pair of orthonormal frames -- as subspaces -- to ``sandwich'' the labeled data via  optimizing for two objectives jointly: i) minimize the distance between the origins of the two subspaces, and ii) to maximize the margin between the hyperplanes and the data, either subspace demanding the data to be in its positive and negative orthant respectively. Our proposed objective however leads to a non-convex optimization problem, to which we resort to Riemannian optimization schemes and derive an efficient conjugate gradient scheme on the Stiefel manifold.

% each hyperplane in the frames acting as discriminative classifiers with the objective of maximizing their margin to the data points. 

% pushing the data away from it in a max-margin setup. One such frame in the pair demands data be far from the origin, while the other one requires the data  the origin; the two frames thus sandwiching the data within a minimal volume. Our proposed objective however leads to a non-convex optimization problem, to which we resort to Riemannian optimization schemes and derive an efficient conjugate gradient scheme on the Stiefel manifold.  

To study the effectiveness of our scheme, we propose a new dataset~\emph{Dash-Cam-Pose}, consisting of clips with skeleton poses of humans seated in a car, the task being to classify the clips as normal or abnormal; the latter is when any human pose is out-of-position with regard to say an airbag deployment. Our experiments on the proposed Dash-Cam-Pose dataset, as well as several other standard anomaly/novelty detection benchmarks demonstrate the benefits of our scheme, achieving state-of-the-art one-class accuracy. 

% dubbed One-class Discriminative Subspace (ODS). Our scheme consists of a pair of subspace classifiers; each  consisting of a set of discriminative hyperplanes which are trained to classify the data points into its half-spaces; the directions in one subspace classifying all data points to its positive half-spaces, while the other to its negative half-spaces. Our objective minimizes the distance between these two subspaces -- thus capturing the minimal volume bounded by the two subspaces containing all the labeled data. Use of ODSP has the advantage that it is richer in its discriminative power, while avoiding any distributional assumptions on the data. We cast the discriminative subspace finding problem as one on the non-linear Stiefel manifold and resort to an efficient Riemannian optimization scheme for the efficient solution. A second contribution of this paper is a novel dataset, Dash-Cam-Pose, with poses extracted from people seated in cars; the goal being to detect if they are seated normally or in poses that are potentially dangerous if airbags deployed. We cast this problem as one of pose anomaly detection and evaluate the adequacy of our ODSP in detecting abnormal poses. We also provide experiments on several standard anomaly detection benchmarks. Our results demonstrate clear benefits of our scheme against classic and recent popular one-class methods.
\end{abstract}

%% file: intro.tex
\section{Introduction}
\label{sec:intro}
There are several real-world problems in which it may be easy to characterize the normal operating behavior of a system or collect data for it, however may be difficult or sometimes even impossible to have data when a system is at fault or is improperly used. Examples include but not limited to an air conditioner making an unwanted vibration, a network attacked by an intruder, abnormal patient conditions such as heart rates, 
an accident captured in a video surveillance camera, or a car engine firing at irregular intervals, among others~\cite{chandala2009anomaly}. In machine learning literature, such problems are usually called one-class problems~\cite{bishop1994novelty,ritter1997outliers}, signifying the fact that we may be able to have unlimited supply of labeled training data for one-class (corresponding to the normal operation of the system), but do not have any labels or training data for situations corresponding to abnormalities. The main goal of such problems is thus to learn a model that fits to the normal set, such that abnormalities can be characterized as outliers of this model.

\begin{figure}[t]
\centering
\subfigure[OC-SVM]{\label{fig:oc-svm}\includegraphics[width=3.5cm,trim={7.5cm 5.5cm 14.5cm 3cm},clip]{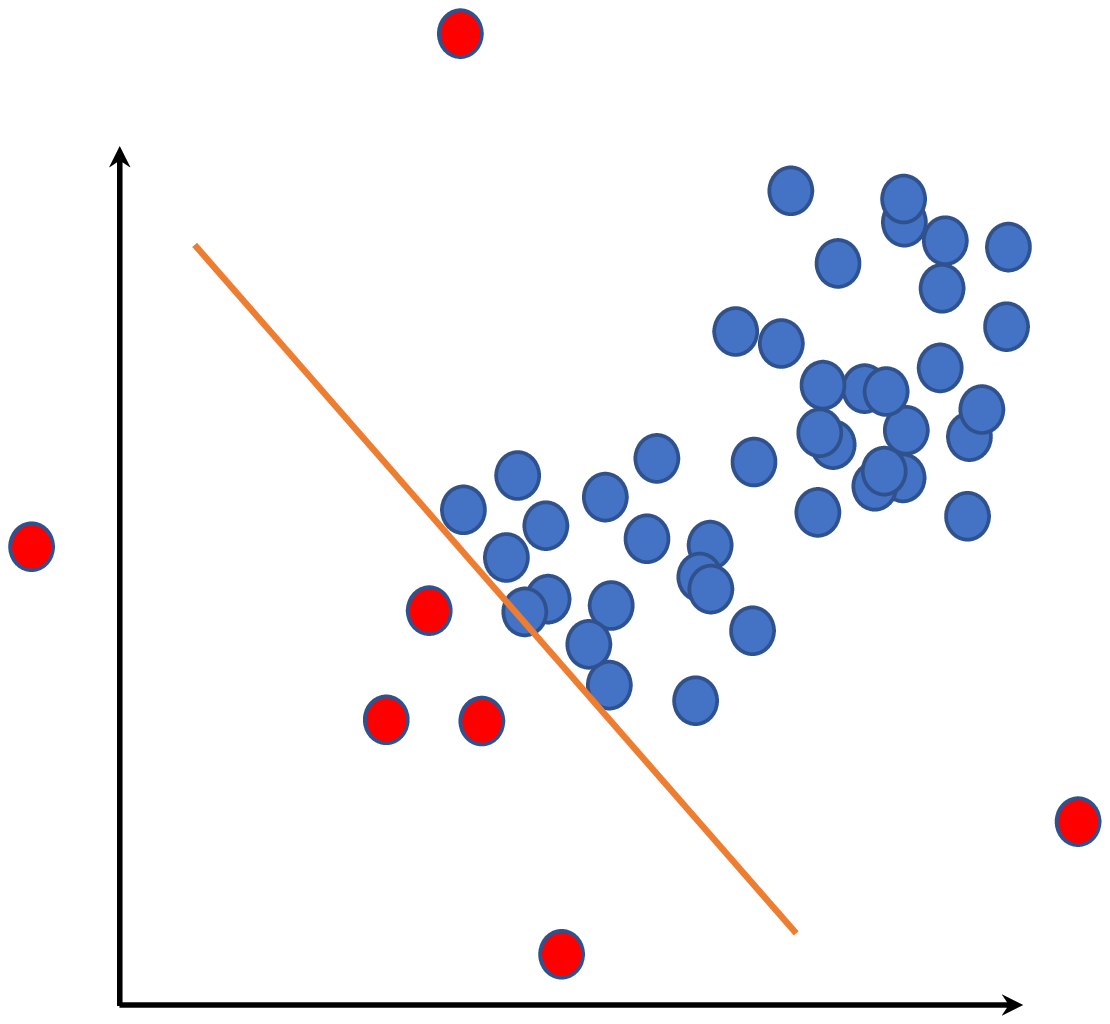}}\hspace*{0.4cm}
\subfigure[SVDD]{\label{fig:svdd}\includegraphics[width=3.5cm,trim={7.5cm 5.5cm 14.5cm 3cm},clip]{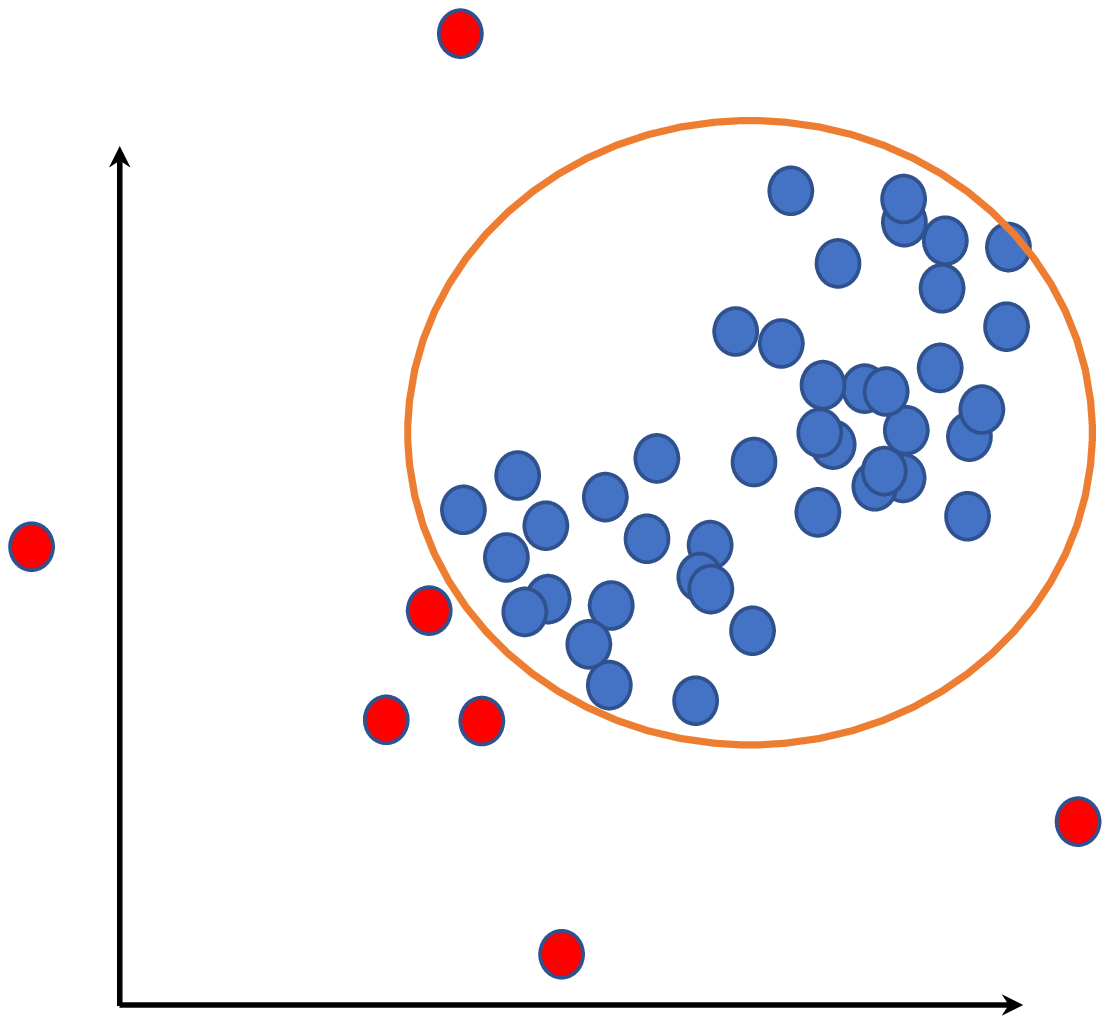}}\hspace*{0.4cm}\\
\subfigure[BODS (ours)]{\label{fig:bods}\includegraphics[width=3.5cm,trim={7.5cm 5.5cm 14.5cm 3cm},clip]{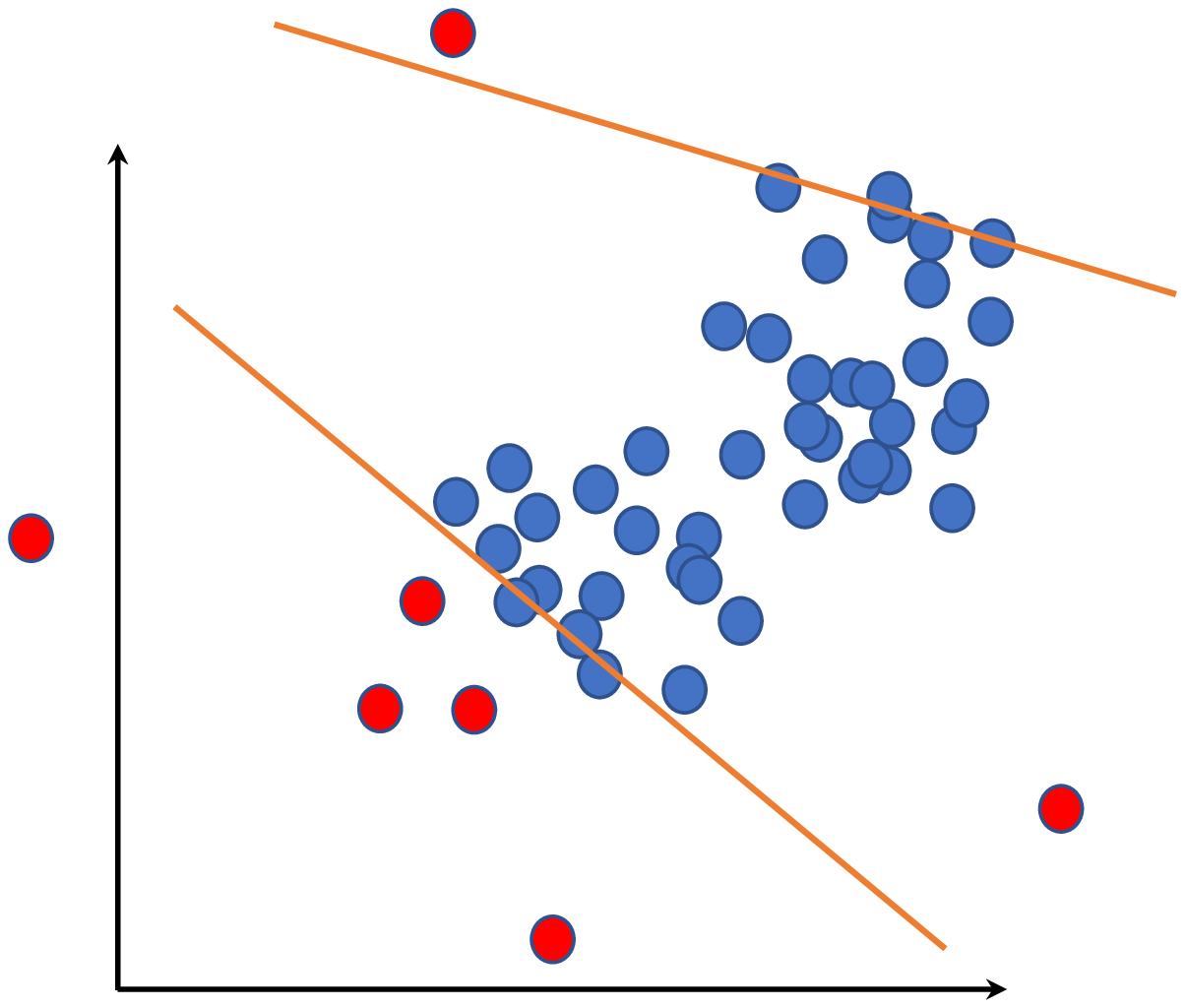}}\hspace*{0.4cm}
\subfigure[GODS (ours)]{\label{fig:gods}\includegraphics[width=3.5cm,trim={7.5cm 5.5cm 14.5cm 3cm},clip]{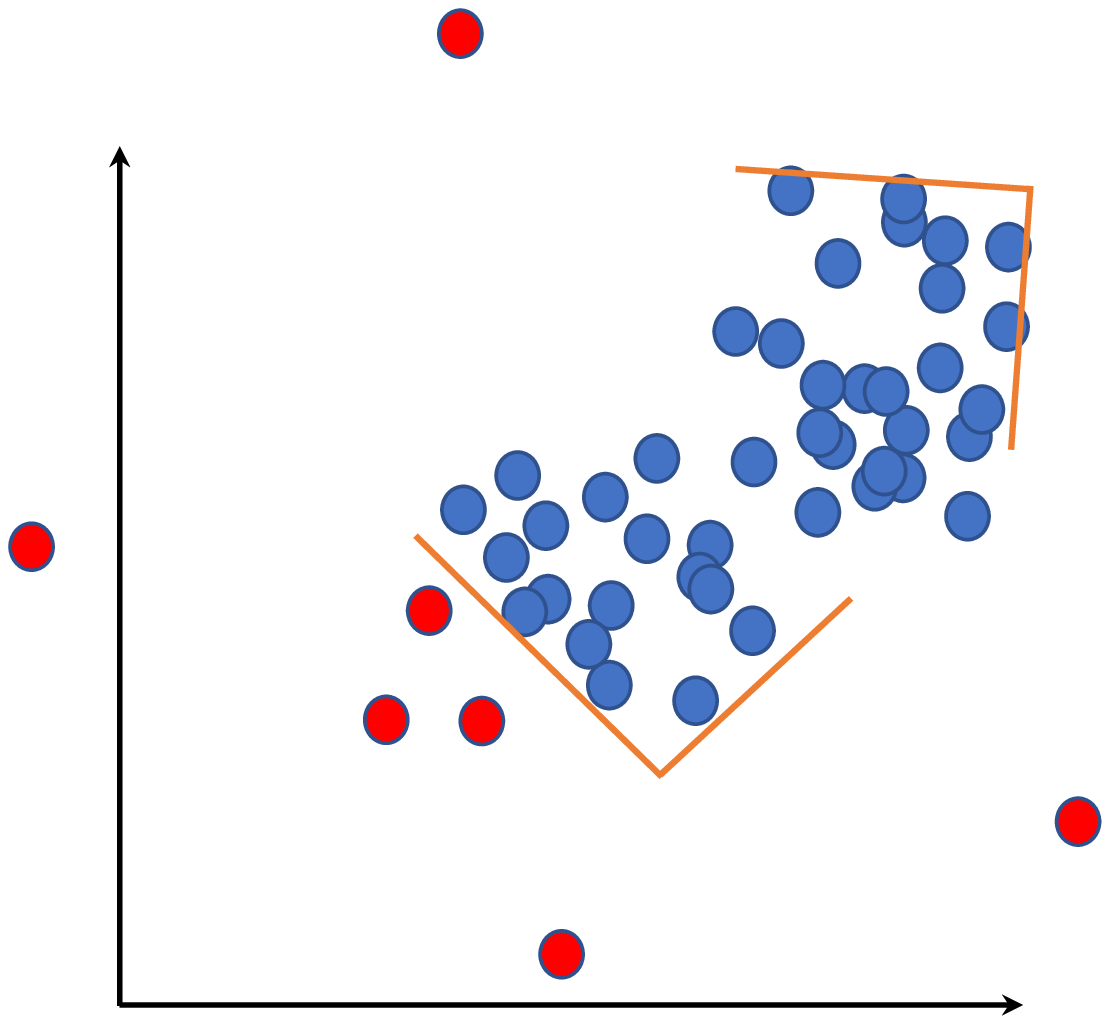}}
\caption{A graphical illustration of classical OC-SVM and SVDD in relation to our proposed BODS and GODS schemes. The blue points show the given one-class data, the red-points are outliers, and decision boundary of each method is shown by orange curves/lines.}
\label{fig:illustration}
\end{figure}

Classical solutions for one-class problems are mainly extensions to support vector machines (SVMs), such as the one-class SVM (OC-SVM) that maximizes the margin of the discriminative hyperplane from the origin~\cite{scholkopf2001estimating}. There are extensions of this scheme, such as the least-squares one-class SVM (LS-OSVM)~\cite{choi2009least} or its online variants~\cite{wang2013online}  that learn to find a tube of minimal diameter that includes all the labeled data. Another popular approach is the support-vector data description (SVDD) that finds a hypersphere of minimum radius that encapsulates the training data~\cite{tax2004support}. There have also been kernelized extensions of these schemes that use the kernel trick to embed the data points in a reproducible kernel Hilbert space, potentially enclosing the `normal' data with arbitrarily-shaped boundaries. 

While, these approaches have shown benefits and have been widely adopted in several applications~\cite{chandala2009anomaly}, they have drawbacks that motivate us to look beyond prior solutions. For example, the OC-SVM uses only a single hyperplane, however using multiple hyperplanes may be beneficial and provide a richer characterization of the labeled set, as also recently advocated in~\cite{wang2018learning}. The SVDD scheme makes a strong assumption on the spherical nature of the data distribution, which may be seldom true in practice. Further, using kernel methods may impact scalability. Motivated by these observations, we propose a novel one-class classification objective that: (i) learns a set of discriminative and orthogonal hyperplanes, as a subspace, to model a multi-linear classifier, (ii) learns a pair of such subspaces, one bounding the data from below and the other one from above, and (iii) minimizes the distances between these subspaces such that the data is captured within a region of minimal volume (as in SVDD). Our framework generates a piecewise linear decision boundary and operates in the input space.

%this may be partially addressed by the least-squares OC-SVM, however, it still uses a single hyperplane to model the data;
Albeit these benefits, our objective is non-convex due to the orthogonality constraints. However, such non-convexity fortunately is not a significant concern as the orthogonality constraints naturally place the optimization objective on the Stiefel manifold~\cite{edelman1998geometry}. This is a well-studied Riemannian manifold~\cite{boothby1986introduction} for which there exist efficient non-linear optimization methods at our disposal. We use one such optimization scheme, dubbed Riemannian conjugate gradient~\cite{absil2009optimization}, which is fast and efficient.

To evaluate the usefulness of our proposed scheme, we apply it to the concrete setting of detecting abnormal or `out-of-position' human poses~\cite{trivedi2007looking,trivedi2004occupant} in cars; specifically, our goal is to detect if the passengers or the driver are seated "out-of-position" (OOP) as captured by an inward looking dashboard camera. This problem is of at most importance in vehicle passenger safety as humans seated OOP may be subject to fatal injuries if the airbags are deployed~\cite{plank1998analytical,marklund2003optimization,khan2008multiphysics}. The problem is even more serious in autonomous cars, which may not (in the future) have any drivers at all to monitor the safety of the passengers. Such OOP human poses include abnormal positions of the face (such as turning back), legs on the dashboard, etc., to name a few. As it may be easy to define what normal seating poses are, while it may be far too difficult to model abnormal ones, we cast this problem in the one-class setting. As there are no public datasets available to study this problem, we propose a novel dataset,~\emph{Dash-Cam-Pose} consisting of nearly 5K short video clips and comprising of nearly a million human poses (extracted using OpenPose~\cite{cao2016realtime}). Each clip is collected from long Internet videos or Hollywood road movies and weakly-annotated with a binary label signifying if passengers are seated correctly or out-of-position for the entire duration of the clip.

We showcase the effectiveness of our approach on the Dash-Cam-Pose dataset, as well as several other popular benchmarks such as UCF-Crime~\cite{sultani2018real}, action recognition datasets such as JHMDB~\cite{jhuang2013towards}, and two standard UCI anomaly datasets. Our experiments demonstrate that our proposed scheme leads to more than 10\% improvement in performance over classical and recent approaches on all the datasets we evaluate on.

Before moving ahead detailing our method, we summarize below the main contributions of this paper:
\begin{enumerate}
\item We first introduce a one-class discriminative subspace (BODS) classifier that uses a pair of hyperplanes.
\item We generalize BODS to use multiple hyperplanes, termed generalized one-class discriminative subspaces (GODS).
\item We propose a new Dash-Cam-Pose dataset for anomalous pose detection of passengers in cars, and
\item We provide experiments on the Dash-Cam-Pose dataset, as well as four other public datasets, demonstrating state-of-the-art performance.
\end{enumerate}

%% file: background.tex
\section{Background and Related Works}
\label{sec:background}
Let $\dataset\subset\reals{d}$ denote the data set consisting of our one class-of-interest and everything outside it, denoted $\dbar$, be the anomaly set. Suppose we are given $n$ data instances $\dl=\set{\vx_1, \vx_2,\cdots, \vx_n} \subset\dataset$. The goal of one-class classifiers is to use $\dl$ to learn a functional $f$ which is positive on $\dataset$ and negative on $\dbar$. Typically, the label of $\dataset$ is assumed $+1$ and that of $\dbar$ as $-1$.

In One-Class Support Vector Machine (OC-SVM)~\cite{scholkopf2001estimating}, $f$ is modeled as an extension of an SVM objective by learning a max-margin hyperplane that separates the origin from the data points in $\dl$. Mathematically, $f$ has the form $\sign(\vw^T\vx+b$), where $(\vw,b)\in\reals{d}\times\reals{1}$ and is learned by minimizing the following objective:
\begin{equation}
\min_{\vw,b,\xi\geq 0} \half\enorm{\vw}^2 - b+C\!\!\sum_{i=1}^n\xi_i,\subjectto \vw^T\vx_i\geq b-\xi_i, \forall \vx_i\in\dl\notag,
%\sum_{i=1}^n  \max\left(0, 1-(\vw^T\vx_i+b) - \xi_i\right)^2 + C\xi_i,
\label{eq:oc-svm}
\end{equation}
where $\xi_i$'s are non-negative slacks, $b$ is the hyperplane intercept, and $C$ is the slack penalty. As a single hyperplane might be insufficient to capture all the non-linearities associated with the one-class, there are extensions using non-linear kernels via the kernel-trick~\cite{scholkopf2001estimating}. However, as is common with kernelized SVM, such a formulation is difficult to scale with the number of data points. 
%SVDD models the one-class functional $f$ as $\sign(\enorm{\vx-\vc}^2 - R^2)$, where $\vc\in\reals{d}$ is the centroid of the hypersphere, and $R$ is its radius, and the optimization
Another popular variant of one-class classifiers is the support vector data description (SVDD)~\cite{tax2004support} that instead of modeling data to belong to an open half-space of $\reals{d}$ (as in OC-SVM), assumes the labeled data inhabits a bounded set; specifically, the optimization seeks the centroid $\vc\in\reals{d}$ of a hypersphere of minimum radius $R>0$ that contains all points in $\dl$. Mathematically, the objective reads:
%These parameters are learned by minimizing the following objective:
\begin{equation}
\min_{\vc,R,\xi\geq 0} \half R^2\!+C\sum_{i=1}^n\xi_i, \subjectto \enorm{\vx_i-\vc}^2\!\leq\!R^2-\xi_i, \forall \vx_i \in \dl,\notag
\label{eq:svdd}
\end{equation}
where, as in OC-SVM, the $\xi$'s model the slack. There have been extensions of this scheme, such as the mSVDD that uses a mixture of such hyperspheres~\cite{lai2015mixture}, density-induced SVDD~\cite{lee2007density}, using kernelized variants~\cite{tax2001one}, and more recently, to use subspaces for data description~\cite{sohrab2018subspace}. A major drawback of SVDD in general is the strong assumption it makes on the isotropic nature of the underlying data distribution. Such a demand is ameliorated by combining OC-SVM with the idea of SVDD in least-squares one-class SVM (LS-OSVM)~\cite{choi2009least} that learns a tube around the discriminative hyperplane that contains the input; however, this scheme also makes strong assumptions on the data distribution (such as being cylindrical). In Figures~\ref{fig:oc-svm} and~\ref{fig:svdd}, we graphically illustrate OC-SVM and SVDD schemes.
% That is,
% \begin{equation}
% \min_{\vw,b,\xi} \half\enorm{\vw}^2 + b + C\sum_{i=1}^n \xi_i^2, \subjectto\vw^T \vx_i + b = \xi_i, \forall \vx_i\in\dl
% \label{eq:ls-ocsvm}
% \end{equation}
%While, the idea of bounding the radius of the one-class tube by the length of the slack is an elegant idea,  There are variants of LS-OCSVM that avoids this problem by turning to kernel methods~\cite{}, or combining with SVDD more explicitly~\AC{\cite{}}.

Unlike OC-SVM that learns a compact data model to enclose as many training samples as possible, a different approach is to use principal component analysis (PCA) (and its kernelized counterpart, such as Kernel PCA and Roboust PCA\cite{candes2011robust,de2003framework,hoffmann2007kernel,nguyen2009robust,xu2013outlier}) to summarize the data by using its principal subspaces. However, such an approach is usually unfavorable due to its high computational cost, especially when the dataset is large. Similar in motivation to the proposed technique, Bodesheim et al.~\cite{bodesheim2013kernel} use null space transform for novelty detection and while Liu et al.~\cite{liu2014unsupervised} optimize a kernel-based max-margin objective for outlier removal and soft label assignment. However, their problem setups are different from ours in that \cite{bodesheim2013kernel} requires multi-class labels in the training data and \cite{liu2014unsupervised} is proposed for unsupervised learning.

In contrast to these prior methods, in this paper, we explore the one-class objective from a very unique perspective; specifically, to use subspaces as in PCA, however instead of approximating the one-class data, these subspaces are aligned in such a way as to bound the data in a piecewise linear manner, via solving a discriminative objective. We first present a simplified variant of this objective by using two different (sets of) hyperplanes, dubbed Basic One-class Discriminative Subspaces (BODS), that can sandwich the labeled data by bounding from different sides; these hyperplanes are independently parameterized and thus can be oriented differently to better fit to the labeled data. Note that there is a similar prior work, termed Slab-SVM~\cite{fragoso2016one}, that learns two hyperplanes for one-class classification. However, these hyperplanes are constrained to have the same slope, which we do not impose in our BODS model, as a result, our model is more general than Slab-SVM. We extend the BODS formulation by using multiple hyperplanes, as a discriminative subspace, which we call Generalized One-class Discriminative Subspaces (GODS); these subspaces provide better support for the one-class data, while also circumscribing the data distribution. The use of such discriminative subspaces has been recently explored in the context of representation learning on videos in Wang and Cherian~\cite{wang2018learning} and Wang et al.~\cite{wang2018video}, however demands a surrogate negative bag of features found via adversarial means. 

\noindent\textbf{Anomaly Detection:} In computer vision, anomaly detection has been explored from several facets and we refer interested readers to excellent surveys provided in ~\cite{chandala2009anomaly,pimentel2014review} on this topic. Here we pickout a few prior works that are related to the experiments we present. To this end, Adam et al.,~\cite{adam2008robust} and Kim et al.~\cite{kim2009observe} use optical flow to capture motion dynamics, characterizing anomalies. A Gaussian mixture modelling of people and object trajectories is used in~\cite{li2013visual,sillito2008semi} for identifying anomalies in video sequences. Saliency is used in~\cite{itti2000saliency,judd2009learning} and detecting out-of-context objects is explored in~\cite{choi2012context, park2012abnormal} using support graph and generative models for characterizing normal and abnormal data. We are also aware of recent deep learning methods for one-class problems.  Feature embeddings (via a CNN) is explored in~\cite{lee2018simple, principled_srikant} minimizing the ``in-distribution'' sample distances, so that ``out-of-distribution'' samples can be found via suitable distance measures. Differently, we attempt at finding a suitable ``in-distribution'' data boundary which is agnostic to the data embedding. A deep variant of SVDD is proposed in~\cite{ruff2018deep}, however assumes the one-class data is unimodal. There are extensions of OC-SVM to a deep setting in~\cite{xu2017detecting,chalapathy2018anomaly,perera2018learning}.  Due to the immense capacity of modern CNN models, it is often found that the learned parameters overfit quickly to the one-class; requiring heuristic workarounds for regularization or avoiding model collapse. Thus, deep methods so far have been primarily used as feature extractors, these features are then used in a traditional one-class formulation, such as in~\cite{chalapathy2018anomaly}. We follow this trend.% in this paper.
%  %Compactness and discriptiveness losses are used in~\cite{perera2018learning} for training neural nets for anomaly detection. While, we also aim towards learning a compact representation for the data, we propose to approach it from a discriminative perspective via learning subspaces in a max-margin setup.
 %Raghavendra et.al also propose OC-NN to learn the one-class classifier inside the neural network by data driven~\cite{chalapathy2018anomaly}. Compared with our schemes, their formulation and problem set-up are totally different.

% is combined with SVDD more explicitly as in~\AC{\cite{}}.

% Such an issue may be addressed by replacing the Euclidean distance by a more generalized Mahalanobis distance, however in that case, estimating the parameters of the data covariance matrix may pose numerical difficulties. In the sequel, we propose a novel non-linear classifier  that combines the max-margin discriminative half-spaces of the OC-SVM formulation and the bounded set assumptions from SVDD in a one-class discriminative subspace (ODS) classifier. 

%In this section, we will exposit our notation and review a few classic approaches for one-class classification on wthehich our proposed scheme is based. 

%% file: algo.tex
\section{Proposed Method}
\label{sec:algo}
 Using the notation above, in this section, we formally introduce our schemes.
 First, we present our basic idea using a pair of hyperplanes, which we generalize using a pair of discriminative subspaces for one-class classification. 

\subsection{Basic One-class Discriminative Subspaces}
Suppose $(\vw_1, b_1)$ and $(\vw_2, b_2)$ define the parameters of a pair of hyperplanes respectively; our goal in the basic variant of one-class discriminative subspace (BODS)  classifiers is to minimize an objective such that all data points $\vx_i$ be classified to the positive half-space of $(\vw_1,b_1)$ and to the negative half-space of $(\vw_2,b_2)$, while also minimizing a suitable distance between the two hyperplanes. Mathematically, BODS can be formulated as solving:
% Without loss of generality, we assume our data points are unit norm and that $\enorm{\vw_1}=\enorm{\vw_2}=1$.
\begin{align}
\label{eq:0}\min_{\substack{(\vw_1,b_1), (\vw_2,b_2),\\ \xi_1,\xi_2, \beta> 0}} \half\enorm{\vw_1}^2\!+\!\half\enorm{\vw_2}^2 &-\!b_1\!-\!b_2\!+\vOmega(\xi_{1i},\xi_{2i})\\
\label{eq:1}\subjectto \left(\vw_1^T\vx_i - b_1\right) \geq \eta &- \xi_{1i}\\
\label{eq:2}\left(\vw_2^T\vx_i - b_2\right) \leq -\eta &+ \xi_{2i} \\ \label{eq:3}\dist\!\left(\left(\vw_i, b_1\right), \left(\vw_2,b_2\right)\right) \leq \beta,& \forall i=1,2,\cdots, n,
\end{align}
where~\eqref{eq:1} constraints the points such that they belong to the positive half-space of $(\vw_1,b_1)$, while~\eqref{eq:2} constraints the points to belong to the negative half-space of $(\vw_2, b_2)$. We use the notation $\vOmega(\xi_{1i},\xi_{2i})=C\sum_{i=1}^n\left(\xi_{1i}+\xi_{2i}\right)$ for the slack regularization and $\eta>0$ specifies a (given) classification margin. The two hyperplanes have their own parameters, however are constrained together by~\eqref{eq:3}, which aims to minimize the distance $\Dist$ between them (by $\beta$). One possibility is to assume $\Dist$ to be the Euclidean distance, i.e.,
$\dist\!\left(\left(\vw_1,b_1\right),\left(\vw_2,b_2\right)\right) = \enorm{\vw_1-\vw_2}^2 + (b_1-b_2)^2$. %\AC{An illustration of this idea is presented in Figure~\ref{fig:bods}}. 

It is often found empirically, especially in a one-class setting, that allowing the weights $\vw_i$'s to be unconstrained leads to overfitting to the labeled data; a practical idea is to explicitly regularize them to have unit norm (and so are the data point $\vx_i$'s), i.e., $\enorm{\vw_1} = \enorm{\vw_2}=1$. In this case, these weights belong to a unit hypersphere $\unitsphere{d-1}$, which is a sub-manifold of the Euclidean manifold $\reals{d}$. Using such manifold constraints, the optimization in~\eqref{eq:0} can be rewritten (using a hinge loss variant for other constraints) as follows, which we term as our \emph{basic one-class discriminative subspace} (BODS) classifier.
\begin{align}
&\label{eq:7}P1:=\!\!\!\min_{\substack{\vw_1,\vw_2\in\unitsphere{d-1}\\ \xi_1,\xi_2\geq 0, b_1,b_2}}\valpha(b_1,b_2) -\!2\vw_1^T\vw_2 + \vOmega(\xi_{1i},\xi_{2i}) \\
&+ \sum_{i} \hinge{\eta\!-\!\left(\vw_1^T\vx_i+b_1\right)-\!\xi_{1i}}\!\!+\!\!\hinge{\eta\!+\!\left(\vw_2^T\vx_i+b_2\right)+\!\xi_{2i}}\!\!\notag,
\end{align}
where using the unit-norm constraints $\dist$ simplifies to $-2\vw_1^T\vw_2 + (b_1-b_2)^2$, and $\valpha(b_1,b_2)=(b_1-b_2)^2-b_1-b_2$.  The notation $\hinge{\ \ }$ stands for the hinge loss. In Figure~\ref{fig:bods}, we illustrate the decision boundaries of BODS model. %In the next section, we generalize the above formulation to use pairs of multiple hyperplanes -- as a subspace -- thus providing a richer and non-linear discriminative setup.

\subsection{Generalized One-class Discriminative Subspaces}
%In this section, first we introduce a direct extension of BODS to the generalized variant, following which we exposit computational simplifications. For example, in a single-class setting, one may loosely think of these hyperplanes as corresponding to the principal directions (as in PCA) of the data distribution.
To set the stage, let us first see what happens if we introduce subspaces instead of hyperplanes in BODS. To this end, let $\mW_1,\mW_2\in\stiefel{K}$ be subspace frames -- that is, matrices of dimensions $d\times K$, each with $K$ columns where each column is orthonormal to the rest; i.e., $\mW_1^T\mW_1=\mW_2^T\mW_2=\eye{K}$, where $\eye{K}$ is the $K\times K$ identity matrix. Such frames belong to the so-called Stiefel manifold, denoted~$\stiefel{K}$, with $K$ $d$-dimensional subspaces. Note that the orthogonality assumption on the $\mW_i$'s is to ensure they capture diverse discriminative directions, leading to better regularization; further also improving their characterization of the data distribution. A direct extension of P1 leads to:
\begin{align}
&P2:=\min_{\substack{\mW\in\stiefel{K},\xi\geq 0, \vb}}\!\!\!\!\!\!\distW(\mW_1,\mW_2)+\valpha(\vb_1, \vb_2)+\vOmega(\xi_{1i},\xi_{2i})\notag\\
&\qquad\label{eq:10}+\sum_i\hinge{\eta-\min(\mW_1^T\vx_i+\vb_1)-\xi_{1i}}^2\\
&\qquad\label{eq:11}+\sum_i\hinge{\eta + \max(\mW_2^T\vx_i+\vb_2)+\xi_{2i}}^2,
\end{align}
where $\DistW$ is a suitable distance between subspaces, and $\vb\in\reals{K}$ is a vector of biases, one for each hyperplane. Note that in~\eqref{eq:10} and~\eqref{eq:11},  unlike BODS, $\mW^T\vx_i+\vb$ is a $K$-dimensional vector. Thus,~\eqref{eq:10} says that the minimum value of this vector should be greater than $\eta$ and~\eqref{eq:11} says that the maximum value of it is less than $-\eta$.

% \begin{figure}
% 	\begin{center}
%         \includegraphics[width=0.65\linewidth,trim={0cm 0cm 0cm 0cm},clip]{figure/hyperplane.eps}
% 	\end{center}
% 	\caption{Illustration of GODS for 2D points, where circle and cross are the learned subspaces containing 2 hyperplanes in each of them. The hyperplanes bound the data points to characterize the data distribution.}
% 	\label{fig:hyperplane}
% \end{figure}

Now, let us take a closer look at the $\DistW(\mW_1,\mW_2)$. Given that $\mW_1,\mW_2$ are subspaces, one standard possibility for a distance is the \emph{Procrustes distance}~\cite{chikuse2012statistics,turaga2008statistical} defined as $\min_{\Pi\in\mathcal{P}_K}\fnorm{\mW_1 - \mW_2\Pi}$, where $\mathcal{P}_K$ is the set of $K\times K$ permutation matrices. However, including such a distance in Problem P2 makes it computationally expensive. To this end, we propose a slightly different variant of this distance which is  much cheaper. Recall that the main motivation to define the distance between the subspaces is so that they sandwich the (one-class) data points to the best possible manner while also catering to the data distribution. Thus, rather than defining a distance between such subspaces, one could also use a measure that minimizes the Euclidean distance of each data point from both the hyperplanes; thereby achieving the same effect. That is, we redefine $\distW$ as:
\begin{equation}
\distW(\mW_1,\mW_2, \vb_1,\vb_2|\vx) = \sum_{j=1}^2\enorm{\mW_j^T\vx+\vb_j}^2,
\label{eq:12}
\end{equation}
where now we minimize the sum of the lengths of each $\vx$ after projecting on to the respective subspaces; thereby pulling both the subspaces closer to the data point. Using this definition of $\distW$, we formulate our \emph{generalized one-class discriminative subspace} (GODS) classifier as:
\begin{align}
&P3:=\label{eq:15}\!\!\!\min_{\substack{\mW\in\stiefel{K}\\ \xi\geq 0, b}} F= \half\sum_{i=1}^n \sum_{j=1}^2\enorm{\mW_j^T\vx_i+\vb_j}^2 + \valpha(\vb_1, \vb_2)\notag\\
&\qquad+\vOmega(\xi_{1i},\xi_{2i})+\frac{\nu}{n}\sum_i\hinge{\eta-\min(\mW_1^T\vx_i+\vb_1)-\xi_{1i}}^2\notag\\
&\qquad+\frac{1}{2n}\sum_i\hinge{\eta + \max(\mW_2^T\vx_i+\vb_2)+\xi_{2i}}^2.
\end{align}
Figure~\ref{fig:gods} depicts the subspaces in GODS model in relation to other methods. As is intuitively clear, using multiple hyperplanes allows richer characterization of the one-class, which is difficult in other schemes.
% A practical difficulty with GODS is that the hinge loss term is per data point; demanding one to iterate over all points during the optimization, that could be slow. Instead if we had a variant for which we could use linear algebra operations directly, that could make the solution faster. By re-arranging the terms in~\eqref{eq:15}, by iterating over the subspaces $K$ subspaces (which is much smaller than $n$), we provide below an approximate variant of GODS formulation, termed A-GODS defined as:
% \begin{align}
% &P4:=\!\!\!\min_{\substack{\mW\in\stiefel{K}\\\xi\geq 0, b}} \half\sum_{j=1}^2\fnorm{\mW_j^T\mX+b_j}^2\!+\!\valpha(b_1,b_2)\!+\!\vOmega(\xi_{1i},\xi_{2i})\notag\\
% &\qquad\qquad+\frac{\nu}{n}\sum_{j=1}^2\hinge{\eta - g_j(\mW_j^T\mX + b_j)-\xi_{ji}}^2.
% \label{eq:20}
% \end{align}
% In the above, we assume $\mX$ to be the matrix with each data point $\vx_i$ as its column, and $g$ is defined as $\min$ and $-\max$ for $\mW_1$ and $\mW_2$ respectively. Note that, A-GODS is approximate in the sense that both schemes find hyperplanes satisfying similar constraints, however, GODS uses all data points (violating the constraints) to find subspaces in every optimization iteration, while A-GODS uses only two points (corresponding to the argmin and argmax of the respective constraints). Given the non-convexity of the formulation, there is no guarantee that the solutions of A-GODS and GODS will be the same, however, A-GODS is usually seen to be 5x faster than GODS, while also converging quickly, and performing on par.

\section{Efficient Optimization}
In contrast to OC-SVM and SVDD, the problem P3 is non-convex due to the orthogonality constraints on $\mW_1$ and $\mW_2$.\footnote{Note that the function $\max(0, \min(z))$ for $z$ in some convex set is also a non-convex function.} However, these constraints naturally impose a geometry to the solution space and in our case, puts the $\mW$'s on the well-known Stiefel manifold~\cite{muirhead2009aspects} -- a Riemannian manifold characterizing the space of all orthogonal frames. There exist several schemes for geometric optimization over Riemannian manifolds (see~\cite{absil2009optimization} for a detailed survey) from which we use the Riemannian conjugate gradient (RCG) scheme in this paper, due to its stable and fast convergence. In the following, we review some essential components of the RCG scheme and provide the necessary formulae for using it to solve our objective.

\subsection{Riemannian Conjugate Gradient}
Recall that the standard (Euclidean) conjugate gradient (CG) method~\cite{absil2009optimization}[Sec.8.3] is a variant of the steepest descent method, however chooses its descent along directions conjugate to previous descent directions with respect to the parameters of the objective. Formally, suppose $F(\mW)$ represents our objective. Then, the CG method uses the following recurrence at the $k$-th iteration:
\begin{equation}
    \mW^{k} = \mW^{k-1} + \lambda^{k-1} \alpha^{k-1},
    \label{eq:recc}
\end{equation}
where $\lambda$ is a suitable step-size (found using line-search) and $\alpha^{k-1}=-\grad F(\mW^{k-1}) + \mu^{k-1} \alpha^{k-2}$, where $\grad F(\mW^{k-1})$ defines the gradient of $F$ at $\mW^{k-1}$ and $\alpha^{k-1}$ is a direction built over the current residual and conjugate to previous descent directions (see~\cite{absil2009optimization}[pp.182])). 

When $\mW$ belongs to a curved Riemannian manifold, we may use the same recurrence, however there are a few important differences from the Euclidean CG case, namely (i) we need to ensure that the updated point $\mW^k$ belongs to the manifold, (ii) there exists efficient vector transports\footnote{This is required for computing $\alpha_{k-1}$ that involves the sum of two terms in potentially different tangent spaces, which would need vector transport for moving between them (see~\cite{absil2009optimization}[pp.182].} for computing $\alpha^{k-1}$, and (iii) the gradient $\grad$ is along tangent spaces to the manifold. For (i) and (ii), we may resort to computationally efficient retractions (using QR factorizations; see~\cite{absil2009optimization}[Ex.4.1.2]) and vector transports~\cite{absil2009optimization}[pp.182], respectively. For (iii), there exist standard ways that take as input a Euclidean gradient of the objective (i.e., assuming no manifold constraints exist), and maps them to the Riemannian gradients~\cite{absil2009optimization}[Chap.3]. Specifically, for the Stiefel manifold, let $\nabla_{\mW} F(\mW)$ define the Euclidean gradient of $F$ (without the manifold constraints), then the Riemannian gradient is given by:
\begin{equation}
    \grad F(\mW) = (\mathbf{I} - \mW\mW^T) \nabla_{\mW} F(\mW).
\end{equation}
The direction $\grad F(\mW)$ corresponds to a curve along the manifold, descending along which ensures the optimization objective is decreased (atleast locally).

% One would need exponential maps for solving (i), which may be expensive. However, there exists cheaper retraction operators that can project the iterant to t
% When working with a manifold, the generic objective $\gamma$ can be re-written as:
% \begin{equation}
%     \min_{\mW} \gamma(\mW) \subjectto \mW^T\mW = \mathbf{I},
% \end{equation}
% where now, the descent directions needs to respect the manifold geometry imposed by the constraints. In this regard, the recurrence in~\eqref{eq:recc} requires three major changes, namely i) it needs to ensure the updated weights are projected back to the manifold, ii) the iterations are along the tangent spaces to the manifold, and iii) there exists a way to do vector transport between tangent spaces. For i) ideally, one would need an exponential map, which may be expensive, instead we may use a cheaper retraction~\cite{absil2009optimization}[Example 2.1.3]. As for iii) there exist well-known methods to approximate the vector transport~\cite{absil2009optimization}[pp.182]. Thus, all we need to specify is how to compute the Riemannian gradient $\grad$. Fortunately, for Stiefel manifold, there exist well-known mechanisms that can take a Euclidean gradient of the objective with respect to the variables, and transform them to Riemannian variants. 
Now, getting back to our one-class objective, all we need to derive to use the RCG, is compute the Euclidean gradients $\nabla_\mW F(\mW)$ of our objective in P3 with regard to the variables $\mW_j$'s; the other variables (such as the biases) are Euclidean and their gradients are straightforward, and the joint objective can be solved via RCG on the product manifold comprising the Cartesian product of the Stiefel and the Euclidean manifolds. Thus, the only non-trivial part is the expression for the Euclidean gradient of our objective with respect to the $\mW$'s,  which is given by: 
% which has been known to converge quickly and has been used in several recent works~\cite{harandi2018dimensionality,cherian2017generalized}. This scheme is a variant of the standard conjugate gradient scheme, however differs in that after each gradient descent, the iterant needs to be projected on to the Stiefel manifold, which typically may be expensive. However, one can use the so-called retraction operators~\cite{absil2012projection} that provide computationally cheap projections, while also have convergence guarantees. Another challenge when using Riemannian descent schemes is the need to compute the Riemannian gradient of the objective, which may be cumbersome as one may need to characterize the geometry of the manifold (curvature) for this. 
\begin{equation}
\frac{\partial F}{\partial \mW_j}\!\!=\!\!\sum_{i=1}^n\vx_i\left(\mW_j^T\vx_i+\vb_1\right)^T\!-\!\mZ_{i^*}\!\!\hinge{\eta-\mW_j^T\vx_{i}-b_j-\xi_{ji}},
\label{eq:21}
\end{equation}
where $i^*=h(\mW_j^T\vx_i+\vb_j)$, $h$ abstracts $\argmin_k$ and $-\argmax_k$ for $\mW_1$ and $\mW_2$ respectively, $i^*$ denotes the selected hyperplane index (out of $K$) and $\mZ_{i^*}$ is a $d\times K$ matrix with all zeros, except $i^*$-th column which is $\vx_i$.
% \begin{align}
% \frac{\partial F}{\partial \mW_2} = &2\sum_i\vx_i(\mW_2^T\vx_i+b_2)^T\nonumber\\
% &+2\sum_{i^{**}}\vx_{i^{**}}\hinge{\eta+\mW_2^T\vx_{i^{**}}+b_2+\xi_{2i}}.
% \label{eq:22}
% \end{align}
% Where $i^{*}$ and $i^{**}$ are the elements that satisfy $\min(\mW_1^T\vx_i+b_1)$ and $\max(\mW_2^T\vx_i+b_2)$ respectively.

\subsection{Initialization}
\label{sec:init}
Due to the non-convexity of our objective, there could be multiple local solutions. To this end, we resort to the following initialization of our optimization variables, which we found to be empirically beneficial. Specifically, we first sort all the data points based on their Euclidean distances from the origin. Next, we gather a suitable number (depending on the number of subspaces) of such sorted points near and far from the origin, compute a singular value decomposition (SVD) of these points, and initialize the GODS subspaces using these orthonormal matrices from the SVD. %Specifically, we use the matrices from the nearest data points for initializing the GODS subspace that demands the points to be classified to its positive orthant, and use the orthonormal matrices from the farthest points to initialize the other GODS subspace. 

\section{One-class Classification}
At test time, suppose we are given $m$ data points, and our task is to classify each of them as belonging to either $\dataset$ or $\dbar$. To this end, we use the learned parameters of our problem P3 as above, and compute the score for each point (using~\eqref{eq:15}).  Next, we use $K$-means clustering (we could also use graph-cut) on these scores with $K=2$. Those points belonging to the cluster with smaller scores are deemed to belong to $\dataset$ and the rest to $\dbar$. 

% Once we minimize the objective and learn the parameters $\mW_1,\mW_2,b_1,b_2$, the next question is how we decide if a given data point $\vx$ belongs to the one-class $\dataset$ or to $\dbar$? Given one data point from the testing set and the learned parameters, we are able to get a $K$-dimensional vector, $L_D$, from~\eqref{eq:19} and~\eqref{eq:20}, which is the distance loss term of one data point corresponding to two subspaces. And then, we repeat this for all testing points and apply unsupervised clustering method, such as $K$-means clustering, to split them into two groups and the group with smaller mean value of  $\enorm{L_D}^2$ is the positive class.

%% file: expts.tex
\section{Experiments}
\label{sec:expts}
In this section, we provide experiments demonstrating the performance of our proposed schemes on several one-class tasks, namely (i) out-of-position human pose detection  using the Dash-Cam-Pose dataset, (ii) human action recognition in videos using the popular JHMDB dataset, (iii) UCF-Crime dataset to find anomalous video events, (iv) discriminating sonar signals from a metal cylinder and a roughly cylindrical rock using the Sonar dataset\footnote{\tiny{\url{https://www.kaggle.com/adx891/sonar-data-set}}}, and (v) abnormality detection in a submersible pump using the Delft pump dataset\footnote{\tiny{\url{http://homepage.tudelft.nl/n9d04/occ/547/oc_547.html}}}. Before proceeding, we first introduce our new Dash-Cam-Pose dataset.
\begin{figure*}[htbp]
\centering
\includegraphics[width=3.3cm, height=2.5cm]{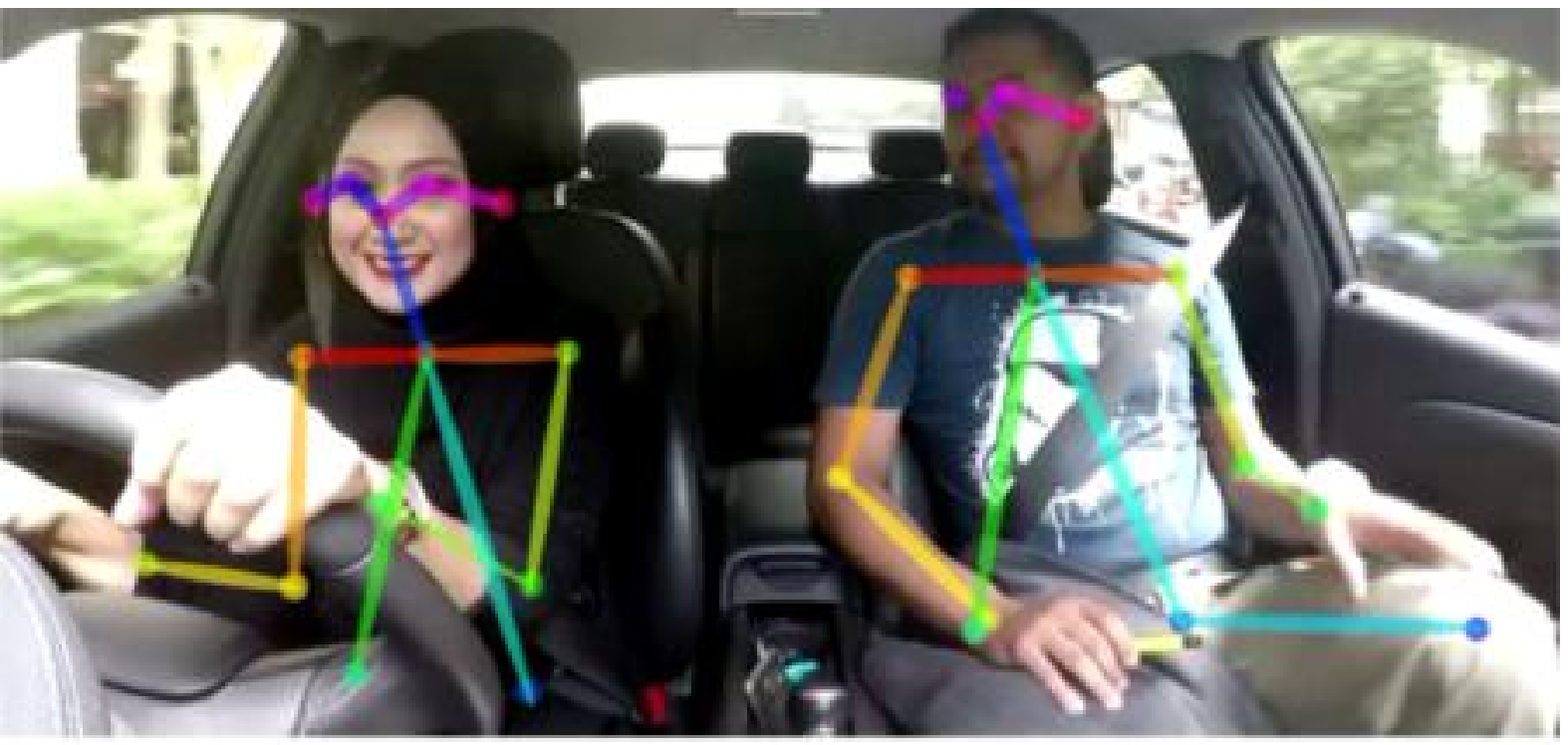}
\includegraphics[width=3.3cm, height=2.5cm]{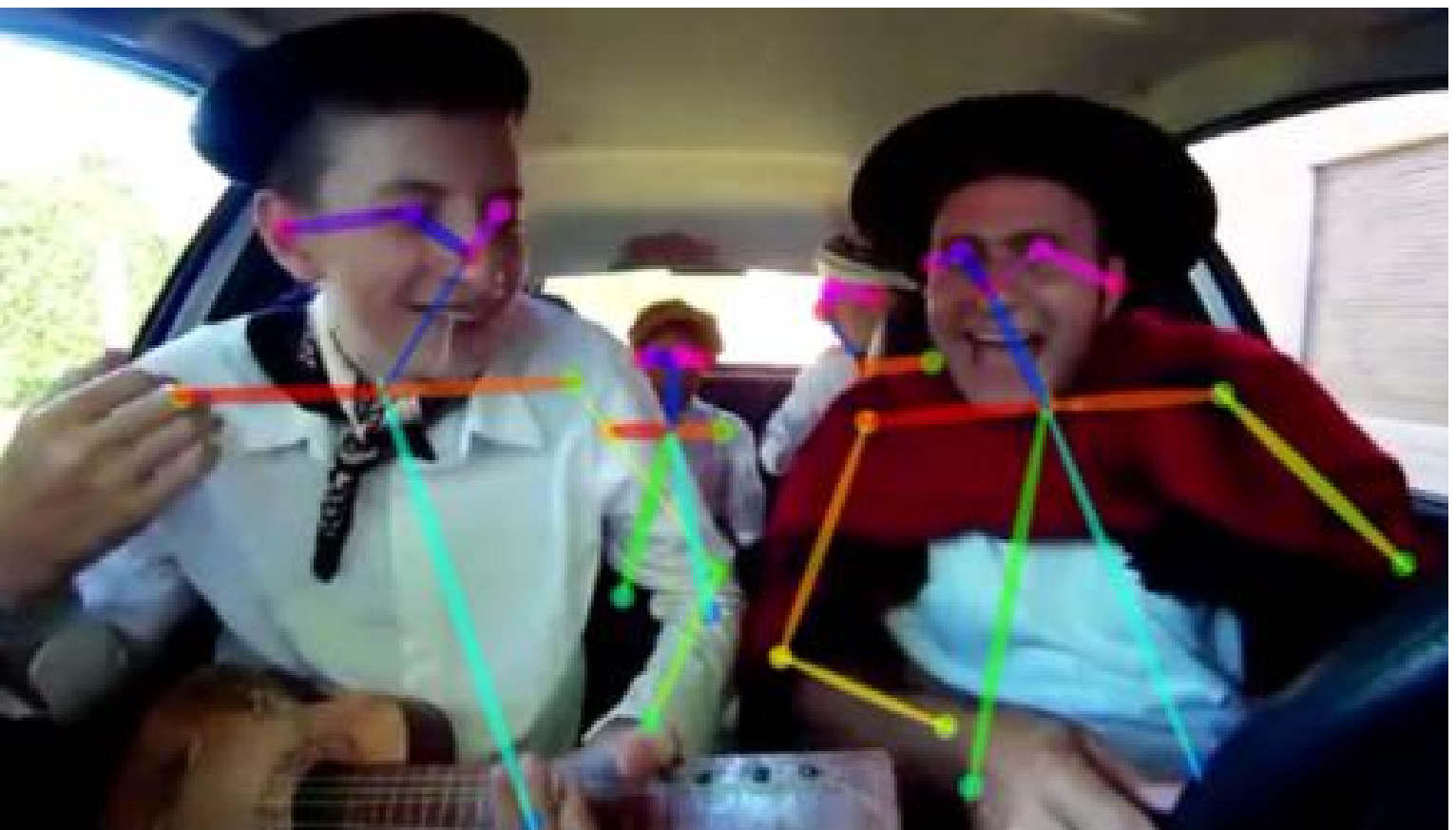}
\includegraphics[width=3.3cm, height=2.5cm]{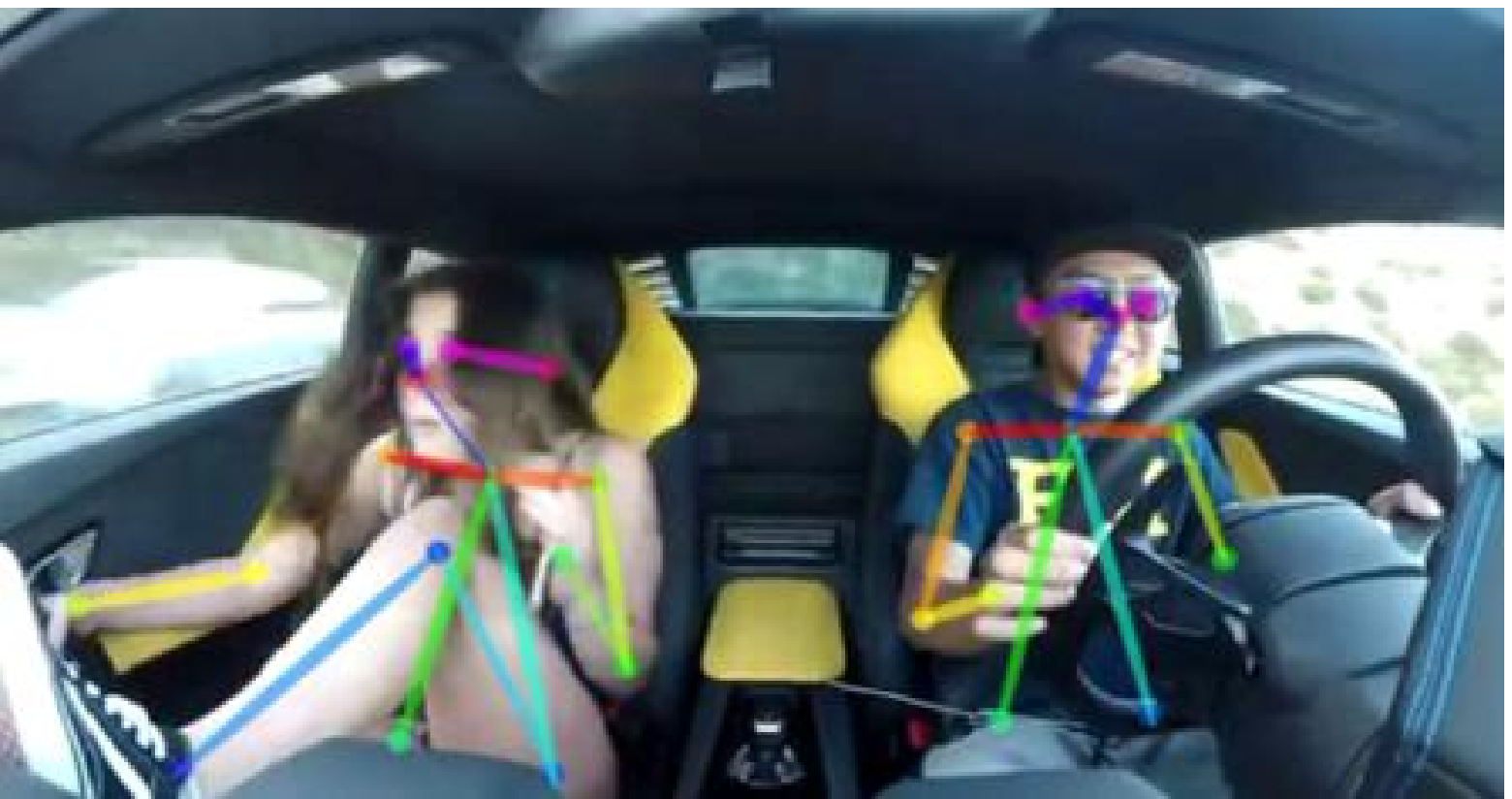}
\includegraphics[width=3.3cm, height=2.5cm]{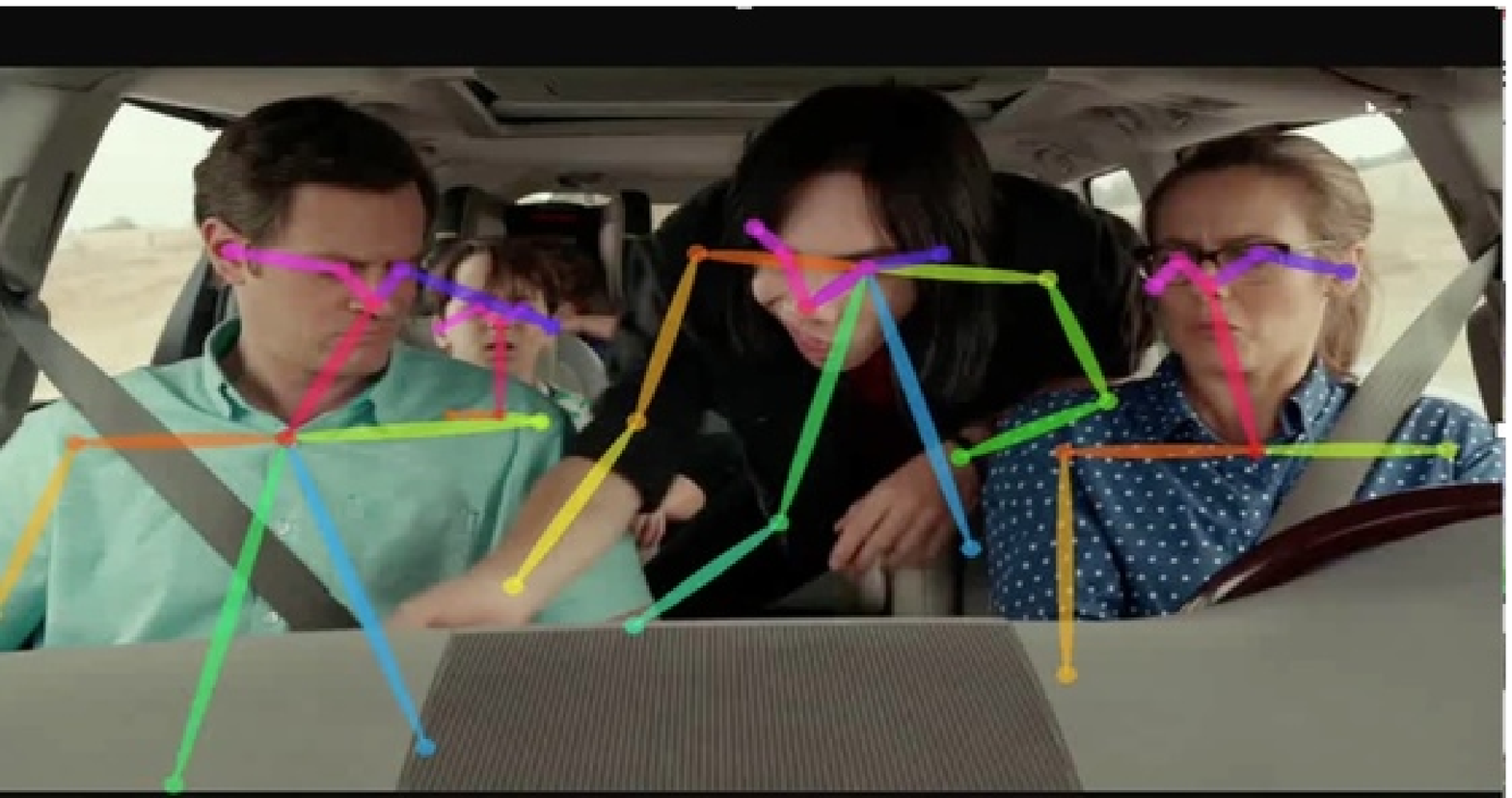}

\caption{Frames from our proposed Dash-Cam-Pose dataset. The leftmost frame has poses in-position (one-class), while the rest of the frames are from videos labeled out-of-position.}
\label{fig:dash-cam-pose}
\end{figure*}

% We implemented our schemes in Matlab using the ManOpt Riemannian optimization package~\cite{boumal2014manopt}.

% In this section, we evaluate our proposed GODS in four datasets with different application scenarios. They are the new Dash-Cam-Pose dataset for anomaly human pose detection task in the car, the JHMDB dataset for image-level anomaly action detection task in the wild environment, Sonar dataset~\cite{gorman1988analysis} for discriminating the sonar signals from a metal cylinder and a roughly cylindrical rock, and Delft pump dataset for detecting abnormal condition of a submersible pump. We will introduce these datasets briefly next along with the detail of the feature we used in each dataset. 
% Tackling this problem has been the main motivation for deriving our one-class schemes.

\subsection{Dash-Cam-Pose: Data Collection}
Out-of-position (OOP) human pose detection is an important problem with regard to the safety of passengers in a car.  While, there are very large public datasets for human pose estimation -- such as the Pose Track~\cite{iqbal2016posetrack} and MPII Pose~\cite{andriluka20142d} datasets, among others -- these datasets are for generic pose estimation tasks, and neither they contain any in-vehicle poses as captured by a dashboard camera, nor are they annotated for pose anomalies. To this end, we collected about 104 videos, each 20-30 min long from the Internet (including Youtube, ShutterStock, and Hollywood road movies). As these videos were originally recorded for diverse reasons, there are significant shifts in camera angles, perspectives, locations of the camera, scene changes, etc. 

To extract as many clips as possible from these videos, we segmented them to three second clips at 30fps, which resulted in approximately 7000 clips. Next, we selected only those clips where the camera is approximately placed on the dashboard looking inwards, which amounted to 4,875 clips. We annotated each clip with a weak binary label based on the poses of humans in the front seat (the back seat passengers often underwent severe occlusions, as a result, was harder to estimate their poses). Specifically, if all the front-seat humans (passengers and the driver) are seated in-position, the clip was given a positive label, while if any human is seated OOP for the entire 3s, the clip was labeled as negative. We do not give annotations for which human is seated in OOP. The in-position and out-of-position criteria are defined loosely based on the case studies in~\cite{nordhoff2005motor,duma1996airbag}, the primary goal being to avoid passenger fatality due to an OOP if airbags are deployed. %To this end, we defined  OOP as poses such as turning back, legs on the dashboard, holding an object near the head, stretching hands out of the windows, bending onto the dashboard, and non-seated passengers. We also classified more than two people in the front seat as an OOP. 

After annotating the clips with binary labels, we applied Open Pose~\cite{cao2016realtime} on each clip extracting a sequence of poses for every person. These sequences are filtered for poses belonging to only the front seat humans. Figure~\ref{fig:dash-cam-pose} shows a few frames from various clips. As is clear from the examples, the OOP poses could be quite arbitrary and difficult to model; which is the primary motivation to seek a one-class solution for this task. In the following section, we detail our data preparation and evaluation scheme. Some statistics of the dataset are provided in Table~\ref{tab:stats}. 

\begin{table}
\centering
\scalebox{0.9}{
\begin{tabular}{l|c}
Dash-Cam-Pose Dataset &\\
\hline
Total \# clips & 4875\\
\% of clips with OOP poses & 28.5\%\\
Total \# poses & 1.06M \\
Total \# OOP poses & 310,996\\
%Total length of videos & 4.06 hrs\\
%Min/Max/Avg. \# poses per frame & 1/11/2.86\\
%Total \# unique videos & 104
\end{tabular}}
\caption{Attributes of the proposed Dash-Cam-Pose dataset.}
\label{tab:stats}
\end{table}

\subsection{Dash-Cam-Pose: Preparation and Evaluation} 
Suitable representation of the poses is important for using them in the one-class task. To this end, we explore two representations, namely (i) a simple bag-of-words (BoW) model of poses learned from the training set, and (ii) using a Temporal Convolutional Network (TCN) ~\cite{kim2017interpretable} which uses residual units with 1D convolutional layers, capturing both local and global information via convolutions for each joint across time. For the former, we 1024 pose centroids, while for the latter the poses from each person in each frame are vectorized and stacked over the temporal dimension. The TCN model we use has been pre-trained on the larger NTU-RGBD dataset~\cite{shahroudy2016ntu} on 3D-skeletons for the task of human action recognition. For each pose thus passed through TCN, we extract features from the last pooling layer, which are 256-D vectors for each clip. 
%To match the expected input of the pre-trained TCN in our 2D pose setup, we map the 2D joint coordinates in our Dash-Cam-Pose dataset with the 3D joints location of NTU-RGBD skeletons by padding zeros for the missing 3rd dimension and joints.

% To make each video descriptor comparable, normalization and exponential embedding are also applied. 

We use a four-fold cross-validation for evaluating on Dash-Cam-Pose. Specifically, we divide the entire dataset into four non-overlapping splits, each split consisting of approximately 1/4-th the dataset, of which roughly 2/3rd's are the labeled positive and the rest are OOP. \emph{We use only the positive data in each split to train our one-class models.} Once the models are trained, we evaluate on the held out split. For every embedded-pose feature, we use the binary classification accuracy against the annotated ground truth for measuring performance. The evaluation is repeated on all the four splits and the performance averaged. %In the next sections, we discuss the other datasets and their evaluation protocols.

\subsection{Public Datasets}
\begin{figure}[t]
\centering
\includegraphics[width=1\linewidth,trim={0cm 0cm 0cm 0cm},clip]{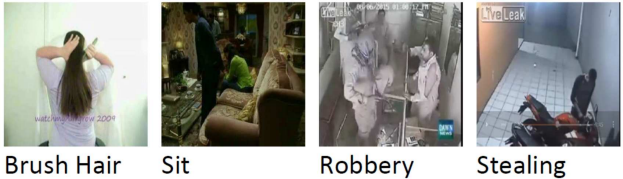}
\caption{Some examples from JHMDB (left-two) and UCF-Crime (right-two) datasets, with respective categories.}
\label{fig:example}
\end{figure}

\noindent\textbf{JHMDB dataset:} is a video action recognition dataset~\cite{jhuang2013towards} consisting of 968 clips with 21 classes (illustrative frames are provided in Figure~\ref{fig:example}). To adapt the dataset for a one-class evaluation, we use a one-versus-rest strategy by  choosing sequences from an action class as ``normal'' while those from the rest 20 classes are treated as ``abnormal''. To evaluate the performance over the entire dataset, we cycle over the 21 classes, and the scores are averaged. For representing the frames, we use an image-net~\cite{krizhevsky2012imagenet} pre-trained VGG-16 model and extract frame-level features from the `fc-6' layer (4096-D). 
%  In details, we follow the original split-1 to decide the train/test videos. After that, the our proposed classifiers are trained on individual frames from training set of videos in one action (recall that we use the RGB stream alone of a standard two-stream CNN model~\cite{simonyan2014two}). During testing, we test the classifier using test videos that may belong to the same action or any other action from the rest of the 20 classes.

\noindent\textbf{UCF-Crime dataset:} is the largest publicly available real-world anomaly detection dataset~\cite{sultani2018real}, consisting of 1900 surveillance videos and 13 categories such as \emph{fighting}, \emph{robbery}, as well as several ``normal'' activities. Illustrative video frames from this dataset and their class labels are shown in Figure~\ref{fig:example}. To encode the videos, we use the state-of-the-art Inflated-3D (I3D) neural network~\cite{carreira2017quo}. Specifically, video frames from non-overlapping sliding windows (8 frames each) is passed through the I3D network; features are extracted from the `Mix\_5c' network layer, that are then reshaped to 2048-D vectors. For anomaly detections on the test set, we first map back the features classified as anomalies by our scheme to the frame-level and apply the official evaluation metrics~\cite{sultani2018real}.

\noindent\textbf{Sonar and Delft pump dataset:} are two UCI datasets, having 208 and 1500 data points respectively, and two classes. We directly adopt the raw feature (60-D and 64-D) without any feature embedding. We keep the train/test ratio as 7/3 while keeping the original proportion of each class in each set. We randomly pick train/test splits and the evaluation is repeated 5 times and performances averaged. %Thus, we create 5 different splits for cross validation and report the averaged performance.

% \begin{figure}[htbp]
% 	\begin{center}
%         \subfigure{\includegraphics[width=0.5\linewidth,trim={0cm 0cm 0cm 0cm},clip]{figure/detect1.eps}}
%         \hspace{-0.6cm}
%         \vspace{-0.8cm}
%         \subfigure{\includegraphics[width=0.5\linewidth,trim={0cm 0cm 0cm 0cm},clip]{figure/detect2.eps}}
%      \subfigure{\includegraphics[width=0.5\linewidth,trim={0cm 0cm 0cm 0cm},clip]{figure/detect3.eps}}
%      \hspace{-0.6cm}
%     \subfigure{\includegraphics[width=0.5\linewidth,trim={0cm 0cm 0cm 0cm},clip]{figure/detect4.eps}}     
%     \vspace{-0.8cm}
% 	\end{center}
% 	\caption{The result of detecting anomalous poses in Dash-Cam-Pose dataset, by backtracking the clip level one-class classification into the human pose tracks. The top two are correct OOPs, while the bottom two are incorrect, the botton-left is a false positives (left) and bottom-right is a false negative.}
%     \label{fig:car-examples}
% \end{figure}

%$F1=\frac{2\times Recall\times Precision}{Recall+Precision}$, Specifically, $Precision=\frac{TP}{TP+FP}$ and $Recall=\frac{TP}{P}$, and the $F1$ score.  

\subsection{Evaluation Metrics}
On the UCF-Crime dataset, we follow the official evaluation protocol, reporting AUC as well as the false alarm rate. For other datasets, we use the F1 score to reflect the sensitivity and accuracy of our classification models. As the datasets we use - especially the Dash-Cam-Pose -- are unbalanced across the two classes, having a single performance metric over the entire dataset may fail to characterize the quality of the discrimination for each class separately, which is of primary importance for the one-class task. To this end, we also report $\text{True Negative Rate}\ TNR=\frac{TN}{N}$, $\text{Negative\ Predictive\ Value}\ NPV=\frac{TN}{TN+FN}$, and $\overline{F1}=\frac{2\times TNR\times NPV}{TNR+NPV}$ alongside standard F1 scores. 

\begin{figure}[htbp]
    \centering
    \subfigure[BODS-Gaussian]{\includegraphics[width=2.5cm,height=2.5cm, trim={6cm 3cm 3cm 1cm},clip]{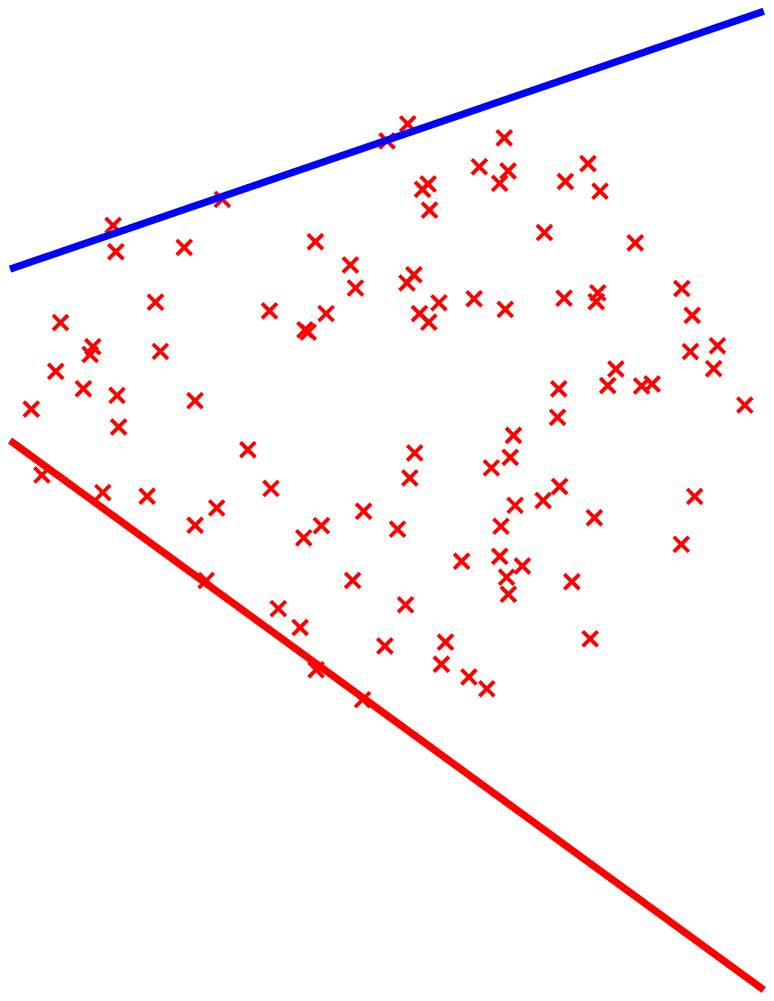}}
    \subfigure[GODS-Gaussian]{\includegraphics[width=2.5cm,height=2.5cm,trim={6cm 3cm 3cm 1cm},clip]{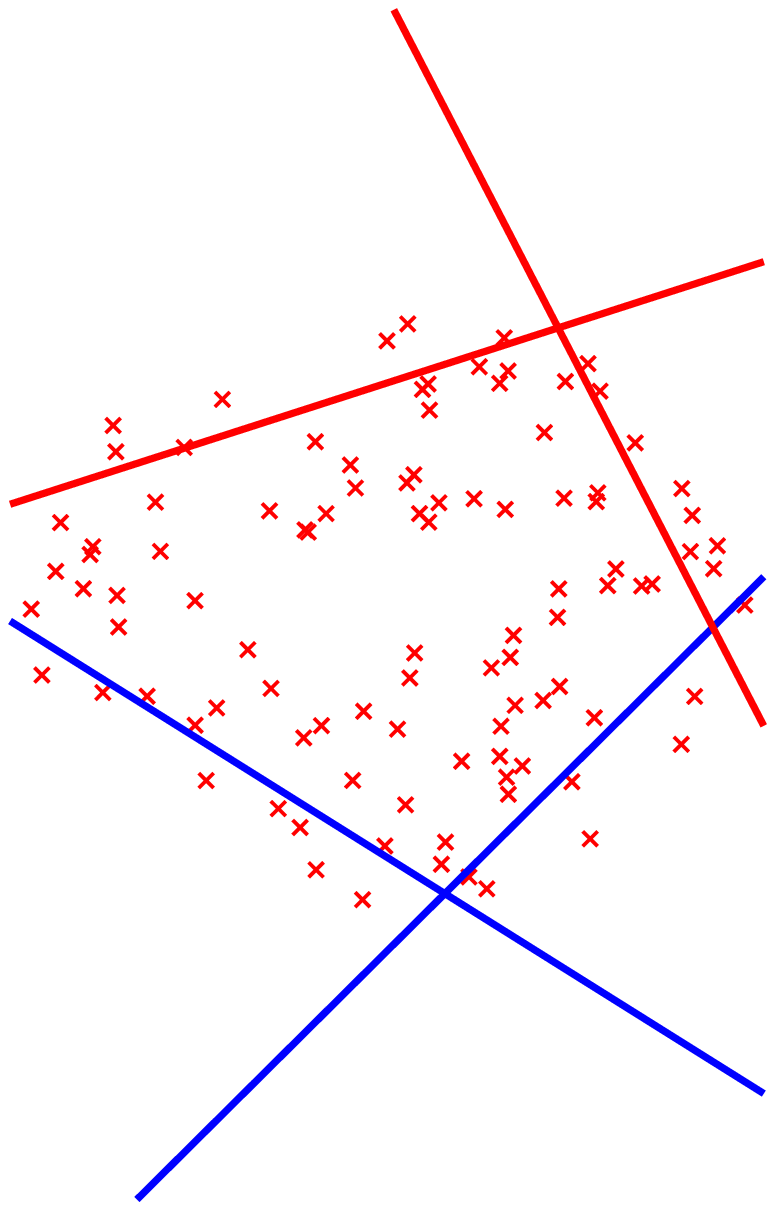}}
    \subfigure[GODS-Arbitrary]{\includegraphics[width=2.5cm,height=2.5cm,trim={6cm 3cm 3cm 1cm},clip]{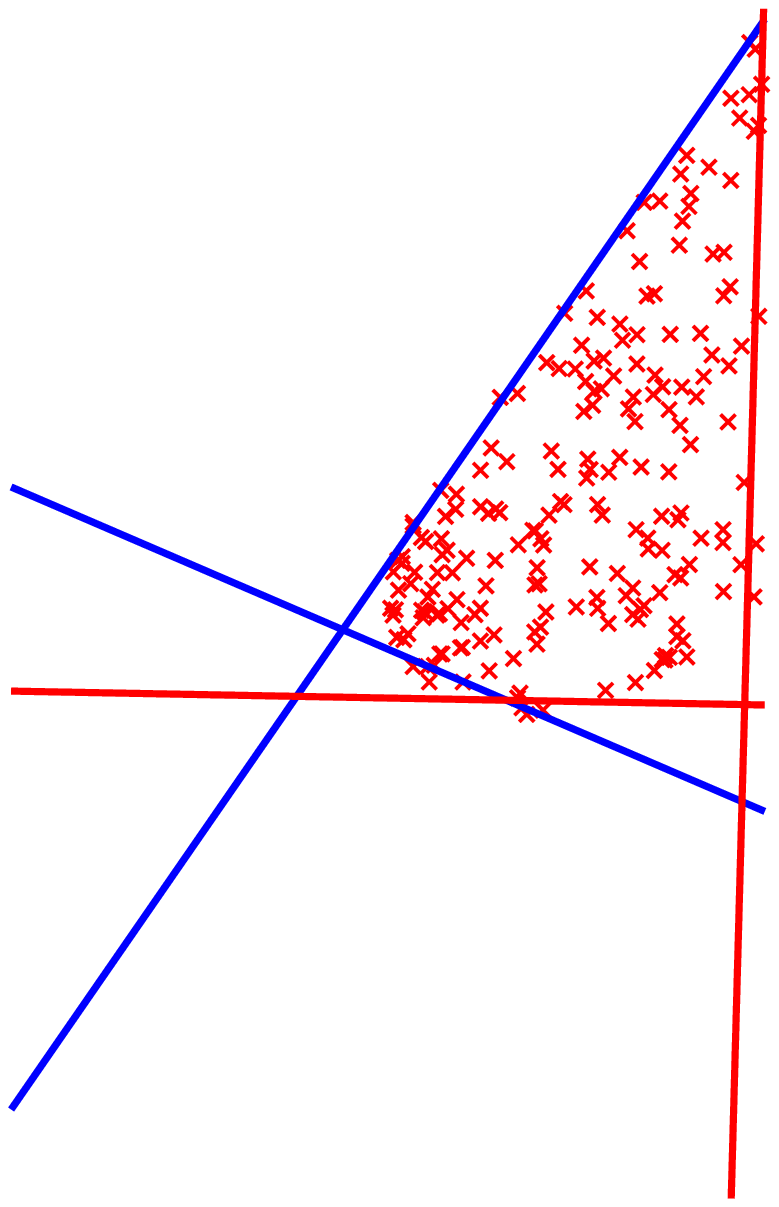}}
\caption{Visualizations of subspaces found by BODS (leftmost) and GODS on various data distributions.}
\label{fig:toy}
\end{figure}
%     \subfigure[GODS-Rectangular]{\includegraphics[width=3cm,height=3cm,trim={10cm 10cm 6cm 10cm},clip]{./figure/gods_visualization_trapezium.pdf}}%\hspace*{1cm}
% As both Dash-Cam-Pose and JHMDB datasets are first introduced under the one-class classification problem, we implement a few popular one-class algorithms as the baseline scheme to compare with. One of the most popular methods is the One-Class Support Vector Machine (OC-SVM)~\cite{scholkopf2001estimating}, which learns a hyperplane that maximize the margin between data and origin. OC-SVM is also the cornerstone of our proposed GODS scheme. Another well-known method is Support Vector Data Description (SVDD)~\cite{tax2004support} that define a hypersphere enclosing the data. Moreover, our proposed GODS classifier is on the non-linear Stiefel manifold, which might be non-comparable with the linear OC-SVM and SVDD. To tackle this problem, we also introduce the non-linear kernelized OC-SVM and SVDD in our experiment for fair comparison, namely K-OC-SVM and K-SVDD respectively\footnote{RBF kernel is used here.}. Besides, we also include the S-SVDD~\cite{sohrab2018subspace} and LS-SVM~\cite{ojeda2008low,choi2009least} as two baseline algorithms, which have been introduced in the Section~\ref{sec:background}.
\begin{figure*}[ht]
	\begin{center}
        \subfigure[UCF-Crime]{\label{subfig:0}\includegraphics[width=0.24\linewidth,trim={0cm 0cm 0cm 0cm},clip]{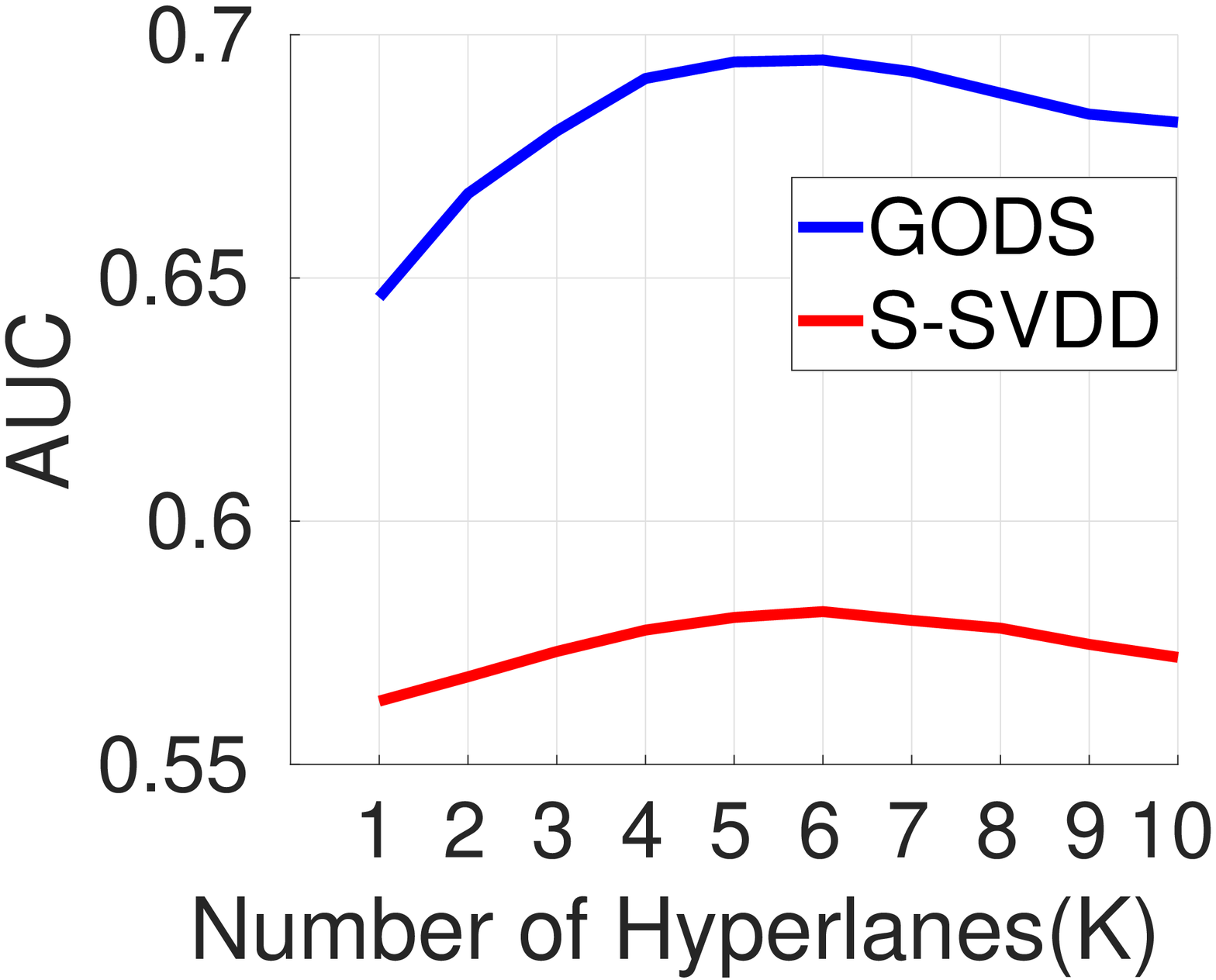}}
        \subfigure[Dash-Cam-Pose]{\label{subfig:1}\includegraphics[width=0.24\linewidth,trim={0cm 0cm 0cm 0cm},clip]{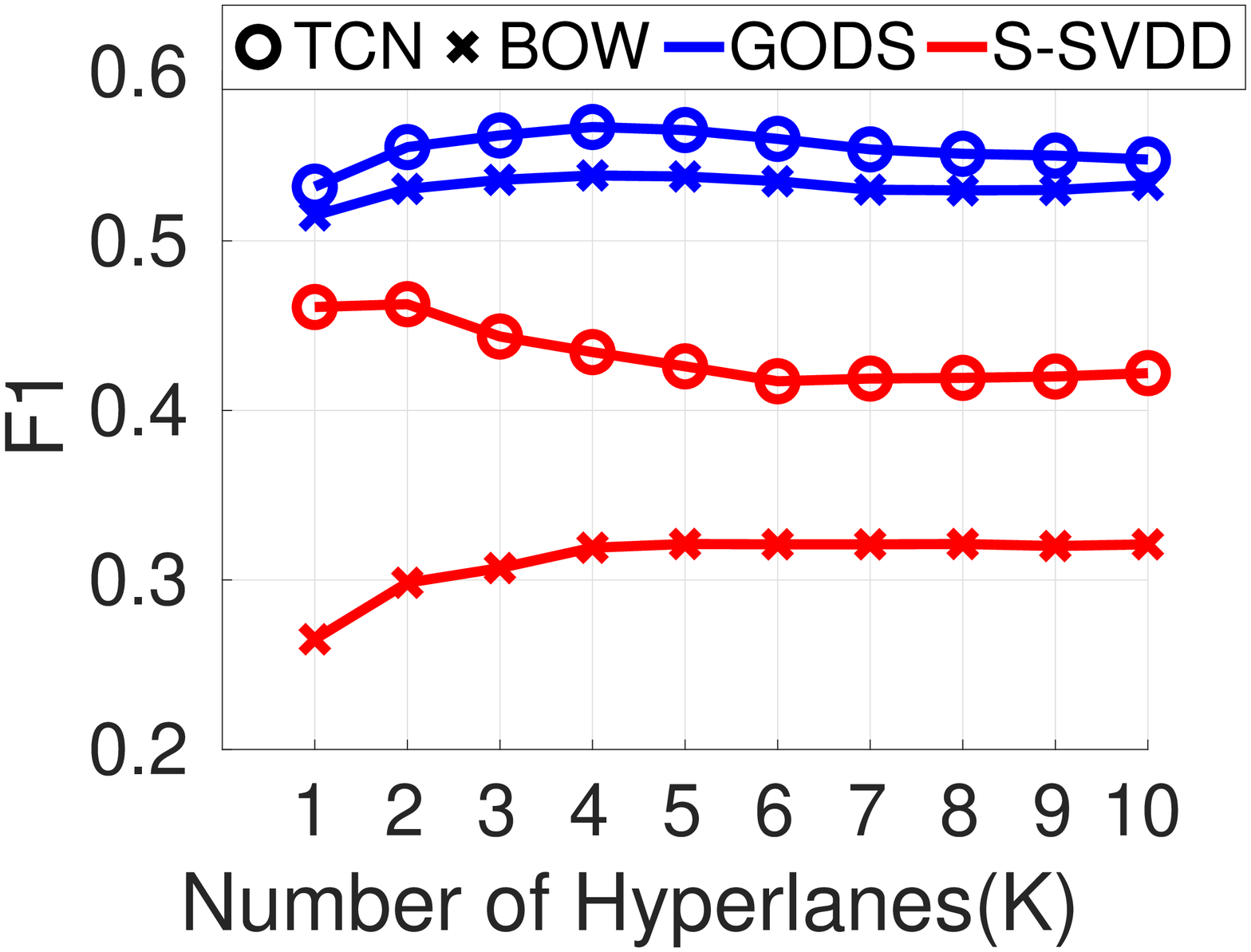}}
        \subfigure[JHMDB]{\label{subfig:2}\includegraphics[width=0.24\linewidth,trim={0cm 0cm 0cm 0cm},clip]{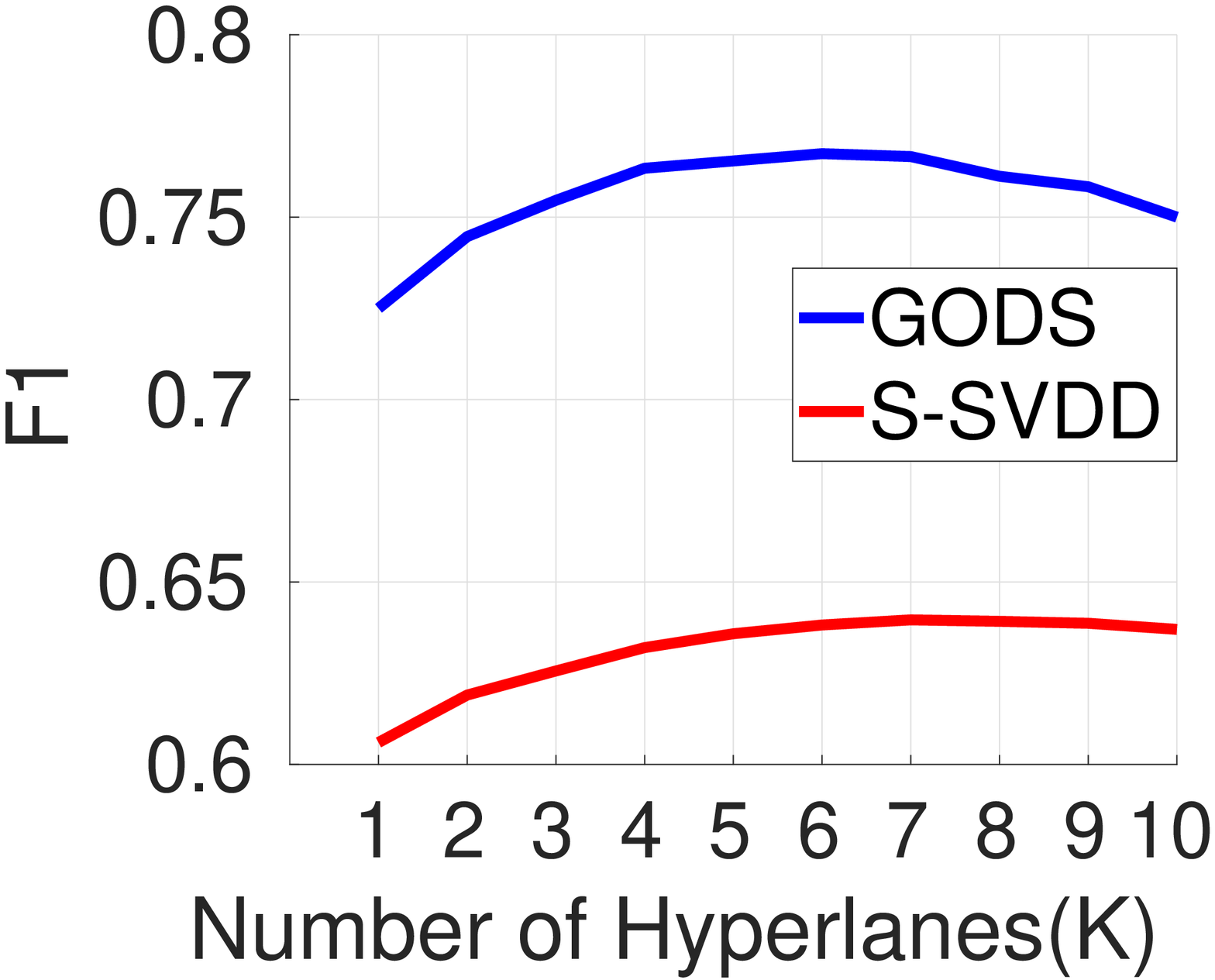}}
     \subfigure[Sonar]{\label{subfig:3}\includegraphics[width=0.24\linewidth,trim={0cm 0cm 0cm 0cm},clip]{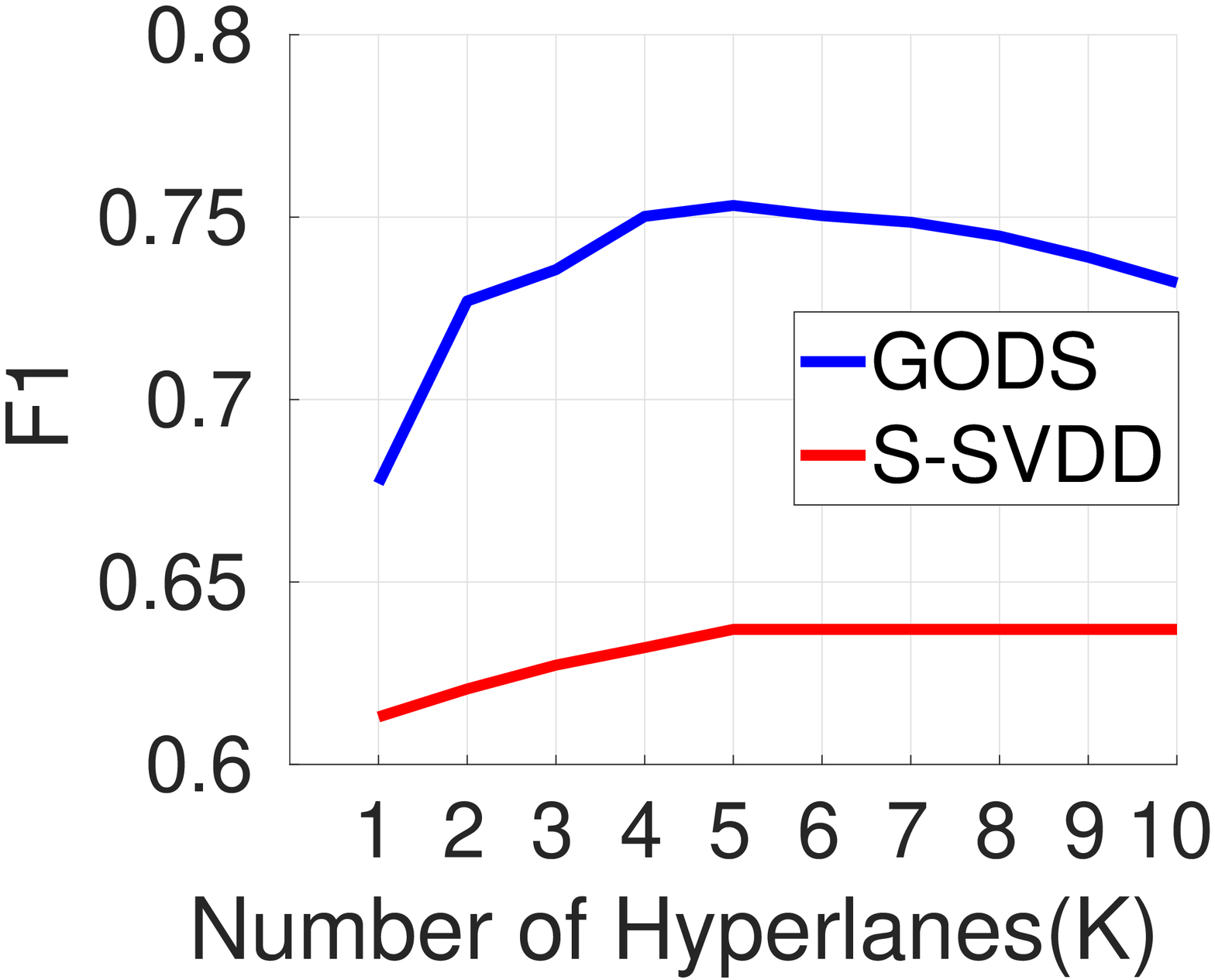}}
    %\subfigure[Pump]{\label{subfig:4}\includegraphics[width=0.19\linewidth,trim={0cm 0cm 0cm 0cm},clip]{figure/pump.eps}}     
	\end{center}
	\vspace*{-0.5cm}
	\caption{Performance of our method on various datasets for an increasing number of subspaces.}
    \label{K_parameter}
\end{figure*}

% \begin{figure}[htbp]
% \centering
% \includegraphics[width=3.3cm, height=2.0cm]{./figure/detect1.jpg}
% \includegraphics[width=3.3cm, height=2.0cm]{./figure/detect2.jpg}
% \includegraphics[width=3.3cm, height=2.0cm]{./figure/detect3.jpg}
% \includegraphics[width=3.3cm, height=2.0cm]{./figure/detect4.jpg}
% \caption{Qualitative results of human poses flagged as OOP by our scheme.}
% \label{fig:car-examples}
% \end{figure}

\begin{figure}[htbp]
    \centering
    \includegraphics[width=0.8\linewidth,clip]{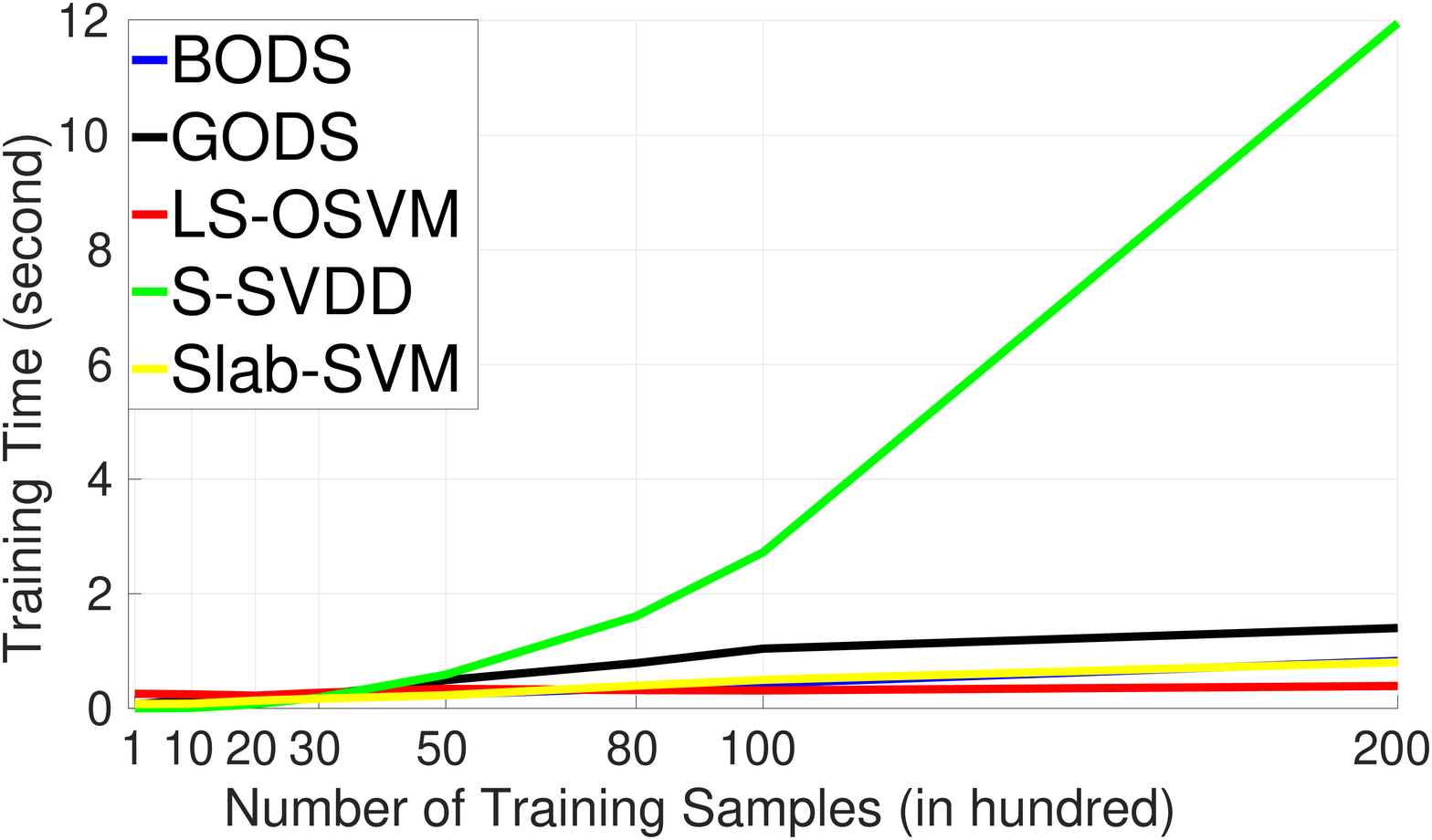}
\caption{Training time of each method with increasing number of training samples.}
\label{fig:training_time}
\end{figure}

%\caption{The result of detecting anomalous poses in Dash-Cam-Pose dataset, by backtracking the clip level one-class classification into the human pose tracks. The top two are correct OOPs, while the bottom two are not, the botton-left is a false positives and bottom-right is a false negative.}
\subsection{Ablative Studies}
\label{exp_result}
%First, we provide a study of the influence of the hyper-parameters in our setup on all the datasets, before furnishing our best results and comparing them to the state of the art.
\noindent\textbf{Synthetic Experiments: } To gain insights into the inner workings of our schemes, we present results on several 2D synthetic toy datasets. In Figure~\ref{fig:toy}, we show four plots with 100 points distributed as (i) Gaussian and (ii) some arbitrary distribution\footnote{The data follows the formula $f(x)=\sqrt{x}*(x+sign(randn)*rand)$, where randn and rand are standard MATLAB functions.}. We show the BODS hyperplanes in the first plot, and the rest two plots show the GODS 2D subspaces with the hyperplanes belonging to each subspace shown in same color. As the plots show, our models are able to orient the subspaces such that they confine the data within a minimal volume. More results  are provided in the supplementary materials.
 \begin{table*}[]
 \centering
 \caption{Average performances on the four datasets, where Dash-Cam-Pose use the $\overline{F1}$ score while the rest use $F1$ score as evaluation metric (classification accuracy is shown in the brackets). K-OC-SVM and K-SVDD are the RBF kernelized variants.}
 \label{soa}
 \scalebox{0.9}{
\begin{tabular}{l|l|l|l|l|l}\hline
Method   & Dash-Cam-Pose\_BOW          & Dash-Cam-Pose\_TCN          & JHMDB                       & Sonar                       & Pump                        \\\hline
OC-SVM~\cite{scholkopf2001estimating}   & 0.167 (0.517)                & 0.279(0.527)                & 0.301 (0.568)                & 0.578 (0.459)                & 0.623(0.482)                \\\hline
SVDD~\cite{tax2004support}     & 0.448 (0.489)                & 0.477(0.482)                & 0.407 (0.566)                & 0.605 (0.479)                & 0.813 (0.516)                \\\hline
K-OC-SVM~\cite{scholkopf2001estimating} & 0.327 (0.495)                & 0.361(0.491)                & 0.562 (0.412)                & 0.565 (0.429)                & 0.601 (0.499)                \\\hline
K-SVDD~\cite{tax2004support}    & 0.476 (0.477)                & 0.489 (0.505)                & 0.209 (0.441)                & 0.585 (0.474)                & 0.809 (0.529)                \\\hline
K-PCA~\cite{hoffmann2007kernel}    &  0.145 (0.502) & 0.258 (0.492) & 0.245 (0.557) & 0.530 (0.426) & 0.611 (0.416)\\\hline
Slab-SVM~\cite{fragoso2016one} & 0.468 (0.568) & 0.498 (0.577) & 0.643 (0.637) & 0.600 (0.619) & 0.809 (0.621)\\\hline
LS-OSVM~\cite{choi2009least}  & 0.234 (0.440)                & 0.246(0.460)                & 0.663(0.582)                & 0.643 (0.466)                & 0.831 (0.448)                \\\hline
S-SVDD~\cite{sohrab2018subspace}   & 0.325 (0.490)           & 0.464 (0.500)           & 0.642 (0.498)           & 0.637 (0.500)           & 0.865 (0.500)          \\\hline\hline
% A-BODS   & 0.471 (0.562)                & 0.512 (0.576)                & 0.665 (0.643)                & 0.635 (0.633)                & 0.828 (0.587)                \\\hline
BODS     & 0.523 (0.582)                & 0.532 (0.579)                & 0.725 (0.714)                & 0.677 (0.662)                & 0.823 (0.714)                \\\hline
%A-GODS   & 0.552 (0.610)  & 0.574 (0.580)  & 0.771 (0.749)         & 0.747 (0.757)           & 0.885 (0.742)  \\\hline
GODS     & \textbf{0.553 (0.629)}          & \textbf{0.584 (0.601) }          & \textbf{0.777 (0.752)} & \textbf{0.762 (0.775) } & \textbf{0.892 (0.755) }        
\end{tabular}}
\end{table*}

\begin{table}[htbp]
\centering
\scalebox{0.9}{
\begin{tabular}{l|c|c}\hline
Method       & AUC & False alarm rate \\\hline
Random       & 50.00                    & -                                    \\\hline
Hasan et al.~\cite{hasan2016learning} & 50.60                    & 27.2                                 \\\hline
Lu et al.~\cite{lu2013abnormal}    & 65.51                   & 3.1                                  \\\hline
$^*$Waqas et al.~\cite{sultani2018real} & 75.41                   & 1.9                                  \\\hline
Sohrab et al.~\cite{sohrab2018subspace} &58.50 &10.5 \\\hline\hline
% A-GODS (K=5)         & 68.85                   & 2.5                                  \\\hline
BODS & 68.26               &2.7      \\\hline           
GODS        & \textbf{70.46}                   & \textbf{2.1}                           
\end{tabular}}
\caption{Performances on UCF-Crime dataset. $^*$Setup is different. }
\label{tab:ucf-soa}
\end{table}

\noindent\textbf{Parameter Study:} In Figure~\ref{K_parameter}, we plot the influence of increasing number of hyperplanes on four of the datasets. We find that after a certain number of hyperplanes, the performance saturates, which is expected, and suggests that more hyperplanes might lead to overfitting to the positive class. We also find that the TCN embedding is significantly better than the BoW model (by nearly 3\%) on the Dash-Cam-Pose dataset when using our proposed methods.  Surprisingly, S-SVDD is found to perform quite inferior against ours; note that this scheme learns a low-dimensional subspace to project the data to (as in PCA), and applies SVDD on this subspace. We believe, these subspaces perhaps are common to the negative points as well that it cannot be suitably discriminated, leading to poor performance. We make a similar observation on the other datasets as well. %, except on the Pump dataset, where S-SVDD performs on par.

% \textbf{Manifold Assumptions and Initializations:} In Table~\ref{tab:ini}, we analyze the influence of two factors, namely (i) if using the Stiefel manifold is useful, against using the Euclidean manifold, and (ii) how well our proposed initialization scheme performs against random initialization or using an average of the data points. The results in Table~\ref{tab:ini} on the JHMDB dataset show that the Stiefel manifold is better (about 2\%) than the Euclidean manifold, and that our initialization scheme is better than other alternatives. 
 
% how Stiefel manifold works in three different initialization methods, where the Euclidean Manifold is used as comparison. Apart from the SVD scheme we introduced in the Section ~\ref{Non-con}, we also apply random Gaussian noise and mean averaged input feature vector as the other two initialization. As is shown in the Figure~\ref{K_parameter}, we use $K=4$ to capture the relatively best result for all datasets. Due to the limited space, we only show the results in JHMDB dataset, while the result in the rest datasets can be found in the supplementary material. As can be seen from the table, Stiefel Manifold benefit the performance in all initialization schemes and the SVD method show the best result.

\subsection{State-of-the-Art Comparisons} In Tables~\ref{soa}, we compare our variants to the state-of-the-art methods. As alluded to earlier, for our Dash-Cam-Pose dataset, as its positive and negative classes are unbalanced, we resort to reporting the $\overline{F1}$ score on the negative set. As is clear from the table, our variants outperform prior methods by a considerable margin. For example, using TCN, GODS is over 30\% better than OC-SVM; even we outperform the kernelized variants by about 20\%. Similarly, on the JHMDB and the other two datasets, GODS is better than the next best method by about 3-13\%, associated with a significant improvement for the classification accuracy (by over 10\%). As the classes used in the test set for these datasets are balanced, we report the F1 scores. Overall, the experiments clearly substantiate the performance benefits afforded by our method on the one-class task. In the Figure~\ref{fig:training_time}, we demonstrate the time consumption for training different models. It can be seen that the GODS \& BODS algorithm are not computationally expensive than other methods, while being is empirically superior (Table~\ref{soa}).
% In Figure~\ref{fig:car-examples}, we show qualitative results on the Dash-Cam-Pose dataset, showing bounding boxes around poses that are flagged as out-of-position by the scheme.

In Table~\ref{tab:ucf-soa}, we present results against the state of the art on the UCF-Crime dataset using the AUC metric and false alarm rates (we use the standard threshold of 50\%). While, our results are lower than~\cite{sultani2018real}, their problem setup is completely different from ours in that they use weakly labeled abnormal videos as well in their training, which we do not use and which as per definition is not a one-class problem. Thus, our results are incomparable to theirs. On other methods for this dataset, our methods are about 5-20\% better.

%% file: conclude.tex
\section{Conclusions}
\label{sec:conclude}
In this paper, we presented a novel one-class learning formulation using subspaces in a discriminative setup, these subspaces are oriented in such a way as to sandwich the data. Due to the non-linear constraints optimization problem that ensues, we cast the objective in Riemannian context however, for which we derived efficient numerical solutions. Experiments on a diverse collection of five datasets, including our new Dash-Cam-Pose dataset, demonstrated the usefulness of our approach achieving state-of-the-art performances.  